\documentclass[runningheads]{llncs}
\usepackage[T1]{fontenc}
\usepackage{graphicx}

\usepackage{subfig}
\usepackage{xcolor}
\usepackage{tcolorbox}
\usepackage{multirow}
\newtcolorbox{graycomment}{
  colback=gray!10,
  colframe=gray!50,
  leftrule=3pt,
  rightrule=0pt,
  toprule=0pt,
  bottomrule=0pt,
  arc=0pt,
  outer arc=0pt,
  left=5pt,
  right=5pt,
  top=3pt,
  bottom=3pt
}

\begin{document}
\title{Mind the XAI Gap: A Human-Centered LLM Framework for Democratizing Explainable AI}

\titlerunning{Mind the XAI Gap}
%
\author{Eva Paraschou\inst{1,3}\orcidID{0000-0002-3561-6994}\thanks{The author was affiliated with the Aristotle University of Thessaloniki when this work was carried out.} \and
Ioannis Arapakis\inst{2}\orcidID{0000-0002-4212-0890} \and
Sofia Yfantidou\inst{1}\orcidID{0000-0002-5629-3493} \and
Sebastian Macaluso\inst{2}\orcidID{0000-0002-4246-4976} \and
Athena Vakali\inst{1}\orcidID{0000-0002-0666-6984}}
\authorrunning{E. Paraschou et al.}
%
\institute{Aristotle University of Thessaloniki, School of Informatics, Thessaloniki, 54124, Greece \\
\email{\{eparascho,syfantid,avakali\}@csd.auth.gr} \and
Telefonica Scientific Research, Barcelona, Spain \\
\email{\{ioannis.arapakis,sebastian.macaluso\}@telefonica.com} \and
Dept. of Applied Mathematics and Computer Science, Technical University of Denmark, 2800 Lyngby, Denmark}

\maketitle              
\begin{abstract}
Artificial Intelligence (AI) is rapidly embedded in critical decision-making systems, however their foundational ``black-box'' models require eXplainable AI (XAI) solutions to enhance transparency, which are mostly oriented to experts, making no sense to non-experts. Alarming evidence about AI's unprecedented human values risks brings forward the imperative need for transparent human-centered XAI solutions. In this work, we introduce a domain-, model-, explanation-agnostic, generalizable and reproducible framework that ensures both transparency and human-centered explanations tailored to the needs of both experts and non-experts. The framework leverages Large Language Models (LLMs) and employs in-context learning to convey domain- and explainability-relevant contextual knowledge into LLMs. Through its structured prompt and system setting, our framework encapsulates in one response explanations understandable by non-experts and technical information to experts, all grounded in domain and explainability principles. To demonstrate the effectiveness of our framework, we establish a ground-truth contextual ``thesaurus'' through a rigorous benchmarking with over 40 data, model, and XAI combinations for an explainable clustering analysis of a well-being scenario. Through a comprehensive quality and human-friendliness evaluation of our framework's explanations, we prove high content quality through strong correlations with ground-truth explanations (Spearman rank correlation=0.92) and improved interpretability and human-friendliness to non-experts through a user study (N=56). Our overall evaluation confirms trust in LLMs as HCXAI enablers, as our framework bridges the above Gaps by delivering (i) high-quality technical explanations aligned with foundational XAI methods and (ii) clear, efficient, and interpretable human-centered explanations for non-experts.

\end{abstract}

\section{Introduction}\label{introduction}
Artificial Intelligence (AI) is becoming increasingly ubiquitous in critical decision-making systems, but its opacity raises major concerns when it comes to safeguarding human ethical values and setting trust and alignment AI boundaries \cite{bellogin2024eu,ozmen2023six}. AI systems' ``black-box'' nature imposes severe transparency risks since their outcomes impact humans who have very limited knowledge about AI systems' underlying reasoning. Such opacity undermines our treasured democratic values and may lead to adverse outcomes, particularly in high-stakes domains, such as healthcare \cite{seyyed2021underdiagnosis,fawzy2022racial}. Regulatory frameworks (e.g., the AI Act\footnote{\url{https://eur-lex.europa.eu/legal-content/EN/TXT/?uri=CELEX\%3A52024AP0138}}) and AI Use solutions (e.g., AI use Taxonomy by NIST\footnote{\url{https://www.nist.gov/publications/ai-use-taxonomy-human-centered-approach}}) urgently call for greater transparency and human-centered explainability in high-risk AI decision-making systems. 
Although rich research is devoted to eXplainable AI (XAI), focus remains primarily on algorithmic transparency, overlooking the varied needs of humans especially across critical domains. It is evident that we must promptly and systematically address the human-AI alignment and safeguard our core democratic values, not only by developing technology-savvy XAI solutions, but by ensuring their actual human-centered focus on openness and interpretability quality. \textbf{Democratizing explainable AI} is a rather challenging endeavour, since human-centered XAI systems must ensure both: (i) transparency, when targeting experts (i.e. practitioners) by revealing their complex inner workings to allow for further evaluation and uptake, and (ii) human-centered explainability, when targeting non-experts (i.e. end-users) to reveal evidence and insights about AI system reasoning in a human-interpretable and comprehensive form \cite{clement2023xair}. Thus, to prioritize human trust and engagement, safeguard human-centered technology uptake and address XAI systems' two-fold role, innovative human-centered XAI (HCXAI) solutions are required  \cite{liao2021human,rozario2023explainable} to bridge the next research and implementation \textbf{\textit{G}}aps in current XAI solutions and lack of human-centered values.

Firstly, the \textbf{``black-box'' nature of the most widely used AI models limits transparency (\textit{G1})}. Most Machine Learning (ML) and Deep Learning (DL) models integrated into AI systems are not inherently transparent and self-interpretable, preventing insights into their inner workings and decision-making processes. However, they remain the preferred choice for many practitioners and researchers due to their effectiveness in specific tasks (e.g., k-means for clustering \cite{geron2022hands}, Convolutional Neural Networks (CNNs) for image classification \cite{lecun2015deep}). In healthcare, even recent studies employ ``black-box'' models (e.g., Random Forests \cite{iwendi2020covid} and CNNs \cite{sultanpure2024hair}), restricting system transparency, an issue now increasingly challenged by emerging regulations in high-stake domains.
%
Over the last decade, XAI methods have significantly enhanced the transparency of ``black-box'' models by providing explanations of their inner workings in various formats (e.g., feature importance, rules) that are understandable by experts developing and deploying them. However, \textbf{XAI methods explanations often make no sense to non-experts (\textit{G2})}. Although practitioners frequently use XAI explanations to interpret AI system decisions to end-users (i.e. non-experts), these explanations are not easily understandable, limiting humans engagement and trust \cite{swamy2023future}. This issue arises because varying levels of technical expertise impose different explainability needs \cite{dhanorkar2021needs}, causing XAI methods to often fall short in practice \cite{burkart2022explainable}. Human-centered explanations are particularly critical in healthcare, as they must be understandable not only to system practitioners who are technical experts, but also to doctors and patients who rely on these systems' decisions.
%
Finally, so far \textbf{human-centered explanations is heavily dependent on post-hoc interpretations (\textit{G3})}. To make XAI method explanations more understandable and human-friendly to non-experts, further post-processing, either by data scientists or Large Language Models (LLMs), is required. Earlier, data scientists and domain specialists applied manual interpretations or visualizations, while more recently, LLMs have been leveraged to further interpret XAI explanations, due to their ability to generate clear and human-friendly outcomes \cite{barua2024concept,ma2023llms}. However, whether involving data scientists, LLMs, or a combination of both, this is a time-consuming process that demands significant human and computational resources, since for new instances, a sequential three-step process is required: model inference, explanation extraction, and subsequent post-processing.

To resolve such critical gaps and democratize explainable AI, we propose a systematic approach that fulfils both experts' transparency and non-experts' human-centered explainability needs. Specifically, we leverage LLMs not merely as interpretation lens but as a means for both extracting and interpreting human-centered explanations within a systematic framework. Building on LLMs’ capabilities as world models, having learned a representation of the world based on vast amounts of data \cite{Gurnee2024language,li2023emergent}, and their ability to excel when provided with additional contextual information (in-context learning) on specific domains and tasks \cite{ouyang2022training,brown2020language}, we explore, for the first time to our knowledge, how this explainability contextual-adaptation can contribute to more effective HCXAI solutions. We trust that the pre-trained knowledge of LLMs enables them, even with limited examples and in-context learning, to understand and contextually adapt to complex explainability principles and tasks (such as computing feature importance), and combined with their excessive content generation capabilities, they serve as a powerful enablers of HCXAI solutions ensuring both transparency and human-centered explainability. Our key contributions are as follows:
\begin{itemize}
\item[\textbf{C1}] \textbf{Establish a benchmark-base as a ground-truth contextual ``thesaurus''}. To further enrich the pre-trained knowledge of LLMs with domain- and explainability-relevant ground-truth, we synthesize a benchmark-base to serve as our contextual ``thesaurus''. To provide domain-relevant ground-truth for LLMs, we perform a clustering analysis on well-being, harvesting knowledge from a rich real-world dataset in the ubiquitous computing domain. Most importantly, for explainability-relevant ground-truth, we apply various foundational XAI methods upon clustering results, which are then rigorously evaluated. Thus, we demonstrate how a foundational ``black-box'' model can enhance its transparency, reveal its inner workings (\textit{\textbf{G1}}) and how non-human-friendly XAI explanations can be transformed into more interpretable, human-centered insights for non-experts (\textit{\textbf{G2}}) through our systematic approach. This extensive ``thesaurus'' benchmarking synthesizes over 40 data, model, and XAI combinations, ensuring high-quality (0.93 LIME fidelity) ground-truth contextual input information to fine-tune LLMs.

\item[\textbf{C2}] \textbf{Propose an in-context and human-centered LLM-based framework}. We leverage LLMs (LLaMA3\footnote{\url{https://github.com/meta-llama/LLaMA3}} and Mistral\footnote{\url{https://github.com/mistralai/mistral-inference}}) and employ in-context learning to enable human-friendly explanations grounded in explainability principles, integrating domain- and explainability-relevant contextual information. Our framework is designed with structured prompts and system configuration to fulfil the HCXAI systems' two-fold role delivered in one response: (i) ``black-box'' models' inner workings, targeting experts to ensure transparency; and (ii) human-centered and -friendly explanations targeting non-experts to ensure enhanced understandability. Our framework overpasses barriers of additional post-interpretations and long sequential processes (\textbf{\textit{G3}}), since with in-context learning domain-specific and explainability demonstrators are integrated into the prompt such that pre-trained LLMs better contextually adapt, providing relevant outcomes for new instances, even with few demonstration examples. Thus, we propose a data- and model-agnostic, generalizable, and reproducible\footnote{Code available here: \url{https://github.com/eparascho/LLMs-for-explanations}} urgently needed HCXAI framework \cite{weber2023beyond}.

\item[\textbf{C3}] \textbf{Assess the quality and human-friendliness of the outcomes of our framework}. To evaluate the transparency of technical explanations provided to experts, we compute structure (i.e. coherence, consistency) and content (i.e. rank correlation, distance) quality of our explanations compared to the ground-truth explanations. Also, to evaluate human-friendliness of our human-centered explanations provided to non-experts, we conduct a user study (N=56) utilizing the User Experience Questionnaire (UEQ) \cite{laugwitz2008construction} and its dimensions about ``pragmatic'' and ``hedonic'' quality. Our findings indicate strong correlation to ground-truth explanations (Spearman rank correlation=0.92 for feature importance rankings) and improved human-friendliness (especially in terms of ease of use, efficiency and clarity). Our overall evaluation confirms our trust in LLMs as HCXAI enablers, as our framework addresses the above \textbf{\textit{G}}aps by providing in one response: (i) high-quality technical explanations of a ``black-box'' ML model that align with explanations from foundational XAI methods, and (ii) human-centered explanations that are clear, efficient and easily interpretable by non-experts.
\end{itemize}

\section{LLMs as explainability enablers}\label{relatedwork}
Our systematic approach builds on existing work at the intersection of LLMs and XAI, particularly studies that use LLMs for interpreting model outcomes and generating explanations. LLMs have been used to perform various tasks while simultaneously interpreting results in human language, using domain information and metadata. For example, GPT-3 was used to develop an explainable LLM-augmented system for depression detection from social media content, incorporating domain-expert criteria via the chain of thought (CoT) technique to provide diagnostic evidence alongside the final diagnosis, which was evaluated through experiments and case studies with domain experts \cite{qin2023read}. Similarly, GPT-4 and OpenLlama were used for explainable financial time series forecasting, integrating historical stock prices, company metadata, and economic news into CoT prompts to generate step-by-step explanations for forecasts, which were evaluated for coherence and comprehensibility \cite{yu2023temporal}.
Moreover, LLMs have widely supported recommender systems in prioritizing human interaction and user satisfaction. For instance, GPT-4 and LLaMA2-7B were used to develop an explainable recommender system that generates explanations by integrating user-item interaction histories, metadata, and textual reviews into prompts, evaluated against well-established benchmarks \cite{ma2024xrec}. Correspondingly, GPT-2 was used to create an uncertainty-aware explainable recommender system that produces natural language explanations using prompting techniques and user/item ID vectors to guide the explanation generation process, evaluated with textual quality metrics (e.g., unique sentence ratio, feature coverage ratio) \cite{peng2024uncertainty}.

While the above works transfer domain knowledge (i.e. historical and application -specific metadata) to LLMs, their pre-trained knowledge can be also enriched on the structure and principles of specific tasks. For example, GPT-4 was used to generate explanations for agent behavior by first distilling the agent's policy into a decision tree and then prompting the LLM with information about this structure, with evaluations showing improved plausibility and reduced hallucination rates in the resulting explanations \cite{zhang2023explaining}. Similarly, GPT-4 and GPT-3.5 were used to generate counterfactual explanations for black-box text classifiers, identifying latent features and minimally editing text to flip model predictions, which were assessed for maintaining semantic similarity with the original input while changing predictions \cite{bhattacharjee2024towards}.
In line with the motivation of our work, the custom GPT x-[plAIn] model was developed to improve the accessibility and interpretability of XAI methods by generating clear summaries of XAI explanations tailored for diverse user groups, including non-experts \cite{mavrepis2024xai}. The model adapts explanations to match each audience's knowledge level and interests, integrating user-specific context and XAI method details into the prompts, which were evaluated through use-case studies, and the findings demonstrate the model's effectiveness in providing audience-specific, easily comprehensible explanations that bridge the gap between complex AI technologies and practical applications.

To summarize, to provide human-centered solutions, the above studies focus on either guiding LLMs through domain-specific knowledge \cite{qin2023read,yu2023temporal,ma2024xrec,peng2024uncertainty} and/or task-specific knowledge \cite{zhang2023explaining,bhattacharjee2024towards} to achieve adaptation or requesting LLMs to explain outcomes in a post-processing phase \cite{mavrepis2024xai} (\textbf{\textit{G3}}). In contrast, our framework contributes to the LLMs and XAI intersection by: (a) transferring XAI foundational principles, methodologies, and demonstration output examples into LLMs through in-context learning, therefore achieving explainability-adaptation along with domain-adaptation, and (b) generating transparency-related explanations to experts and human-centered, easily interpretable explanations to non-experts in one single LLM response (\textbf{C2}).

\section{Contextual setting}\label{experimentalsetup}
\begin{figure}[htp]
  \centering
  \includegraphics[width=\textwidth]{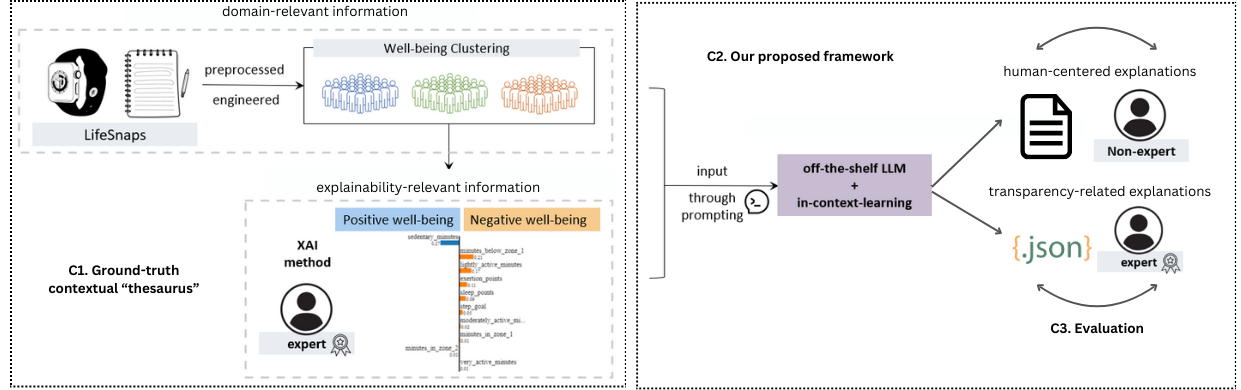}
  \caption{The overview of our work.}
  \label{fig:teaser}
\end{figure}

Our systematic approach builds upon the ground-truth contextual ``thesaurus'' to develop the proposed in-context and human-centered framework, as illustrated in Fig. \ref{fig:teaser}. The ``thesaurus'' serves as a benchmark-base to enrich the domain- and explainability-relevant knowledge of the pre-trained LLMs (\textbf{C1}), as detailed in section \ref{thesaurus}. Using these contextually adapted LLMs, we develop our systematic framework (\textbf{C2}) —an urgently needed HCXAI solution— described in section \ref{methodology}. This framework is then rigorously evaluated across multiple dimensions (\textbf{C3}), as presented in section \ref{evaluation}.

\subsection{Application Domain \& Data}\label{applicationdomain}
As mentioned in section \ref{introduction}, we demonstrate the applicability of our framework in the critical healthcare domain, where democratizing XAI is essential not only for compliance with current legal standards but also to ensure that non-experts (i.e. doctors and patients) receive human-centered explanations, thereby increasing engagement and trust. We focus on the contemporary issue of well-being monitoring, particularly given the growing global prevalence of mental health conditions, with depression predicted to become the leading cause of disease burden by 2030 \cite{funk2016global}. We utilize the LifeSnaps dataset \cite{yfantidou2022lifesnaps}, a publicly available multi-modal dataset in the ubiquitous computing domain, collected in-the-wild from 71 participants across four countries: Sweden, Italy, Greece, and Cyprus. It comprises over 71 million passively acquired data points, including detailed information on sleeping patterns, heart rate measurements, physical activity, and stress levels and additionally, it contains self-reported data on mood, step goals, demographic details, and responses to established surveys such as PAR-Q \cite{warburton2011evidence}, IPIP \cite{goldberg1992development}, PANAS \cite{watson1988development}, and S-STAI \cite{Spielberger1999MeasuringAA}.

Since the LifeSnaps dataset lacks labeled ground-truth information for well-being, we approach well-being monitoring through unsupervised clustering analysis, leveraging the dataset's heterogeneity and cross-modality. We categorize data modalities into seven areas: Physical Activity, Sleep, Health, Mental Health, Demographics, Personality, and Behavior. Following a similar approach to \cite{katevas2018}, we divide the features into two subsets: the training set, used exclusively for deriving clusters and consisting of features that reflect habits and trends, which indicate behavioral patterns that individuals can adjust to improve well-being; and the validation set, which is held out for validity assessment and interpretation of clustering outputs, consisting of 41 indicators of overall well-being. Additionally, we implement extensive preprocessing, including missing values imputation, data normalization, feature engineering, and dimensionality reduction (with further details in appendix \ref{preprocessingappendix}), and after preprocessing and engineering, we identify six feature subsets (detailed in appendix \ref{datasetvariants}) that will be used in the clustering analysis experimentation.


\subsection{The ground-truth contextual ``thesaurus''}\label{thesaurus}
Our proposed benchmark-base ground-truth contextual ``thesaurus'' aims to enrich the pre-trained knowledge of LLMs with domain- and explainability-relevant knowledge (\textbf{C1}), enabling our framework to provide both transparency -related explanations to experts and human-centered, easily interpretable explanations to non-experts. To provide domain-relevant ground-truth, we perform a clustering analysis on well-being using the rich LifeSnaps dataset, experimenting with the following foundational clustering algorithms: k-means \cite{macqueen1967some}, spectral \cite{shi2000normalized}, fuzzy c-means \cite{bezdek1984fcm}, DBSCAN \cite{ester1996density}, HDBSCAN \cite{campello2013density}, and Robust Border Peeling \cite{du2021robp}. We also perform hyperparameter fine-tuning for each algorithm to ensure optimal performance and meaningful clusters. Specifically, for parametric clustering algorithms (i.e. k-means, fuzzy c-means), we employ the elbow method and silhouette score, while for non-parametric algorithms (i.e. DBSCAN, HDBSCAN), we conduct a grid search on key parameters, such as eps, min\_samples, and min\_cluster\_size\footnote{\url{https://hdbscan.readthedocs.io/en/latest/parameter\_selection.html}}. Finally, to assess clustering validity in the absence of ground-truth knowledge, we use the following metrics: Silhouette Score \cite{rousseeuw1987silhouettes}, Davies-Bouldin Index \cite{davies1979cluster}, Calinski-Harabasz Index \cite{calinski1974dendrite}, Dunn Index \cite{dunn1974well}, PBM Index \cite{pakhira2004validity}, and Xie-Beni Index \cite{xie1991validity}.

\begin{figure}[htp]
  \centering
  \includegraphics[width=0.5\textwidth]{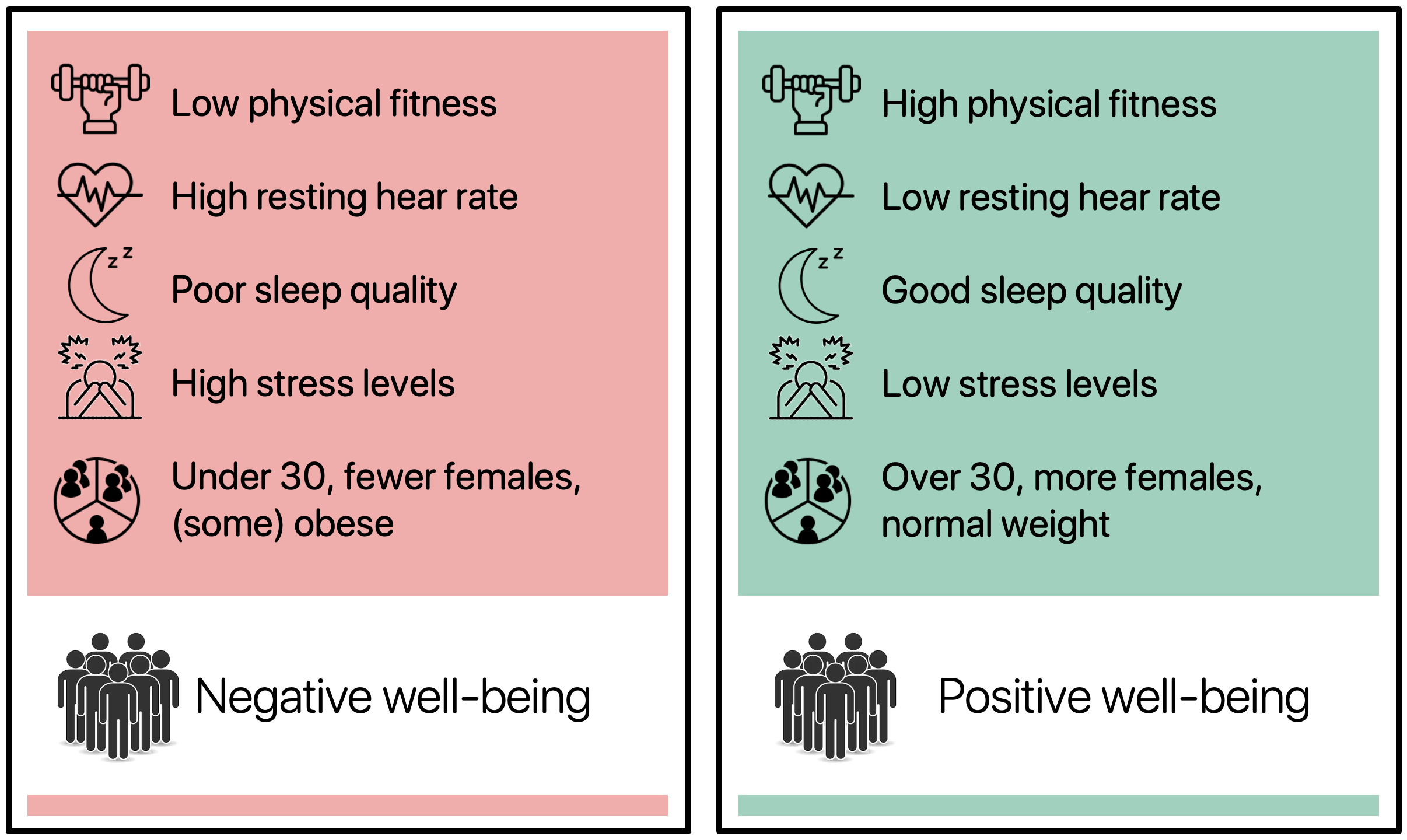}
  \caption{Overview of the individual's characteristics in the two clusters -negative and positive well-being- identified by k-means, achieving 0.4 silhouette score.}
  \label{fig:clusters}
\end{figure}

The best-performing algorithm is k-means, trained on hourly features, such as physical activity, sleep patterns, and health metrics, achieving a silhouette score of 0.35. Given k-means' sensitivity to outliers, which can bias cluster centroids, we remove outliers using the Interquartile Range method, particularly suitable for skewed data distributions \cite{domanski2020study}, improving the silhouette score to 0.4. Detailed results of each clustering algorithm, feature subset, and evaluation metric are presented in appendix \ref{appandixB}. The k-means algorithm revealed two primary clusters, which we characterize based on their well-being status using the validation set features. To do so, we analyze the distribution of these features, assess their statistically significant differences ($p<0.05$) between the two clusters using the Mann-Whitney test \cite{nachar2008mann}, and finally assign labels to the clusters (see Fig. \ref{fig:clusters}).


The need for human-centered explanations is clear, as, despite the valuable population-level insights provided by the clustering analysis on individual well-being, because of its ``black-box'' nature, it remains unclear why certain individuals at specific data points exhibit particular well-being characteristics, which could discourage further user engagement (\textbf{\textit{G1}}). Therefore, to provide explainability-relevant ground-truth, we apply multiple foundational XAI methods to the clustering results to identify the most discriminatory features that informed the cluster assignments for individual data points and generate relevant explanations. We experiment with the following four XAI methods, covering different combinations of key attributes (i.e., local vs. global, numeric vs. rule-based, feature-importance-based vs. example-based), along with their suggested evaluation metrics: i) LIME \cite{ribeiro2016should} for local feature-importance-based explanations, evaluated using the fidelity metric \cite{molnar2020interpretable}; ii) Coefficients for global white-box explanations, evaluated using accuracy and F1; iii) Anchor \cite{ribeiro2018anchors} for local rule-based explanations, evaluated using the coverage and precision metrics \cite{molnar2020interpretable}; and iv) Counterfactuals \cite{wachter2017counterfactual} for local example-based explanations, evaluated using proximity and sparsity metrics.

\begin{figure}[ht!]
    \centering
    \begin{minipage}{0.45\textwidth}
        \centering
        \tiny
        \begin{tabular}{r|cc}
        \hline
        \multicolumn{1}{c|}{\textbf{Method}} & \multicolumn{2}{c}{\textbf{Metrics}} \\ \hline \hline
        \multirow{2}{*}{Coefficients} & \multicolumn{1}{c|}{\textit{\textbf{Accuracy}}} & \textbf{F1} \\ \cline{2-3} & \multicolumn{1}{c|}{0.99}  & 0.99  \\ \hline \hline
        \multirow{2}{*}{Anchors} & \multicolumn{1}{c|}{\textit{\textbf{Coverage}}}  & \multicolumn{1}{l}{\textit{\textbf{Precision}}} \\ \cline{2-3}  & \multicolumn{1}{c|}{0.23} & 0.98 \\ \hline \hline
        \multirow{2}{*}{LIME} & \multicolumn{2}{c}{\textit{\textbf{Fidelity}}} \\ \cline{2-3} & \multicolumn{2}{c}{0.93} \\ \hline \hline
        \multirow{2}{*}{Counterfactuals} & \multicolumn{1}{c|}{\textit{\textbf{Proximity}}} & \multicolumn{1}{l}{\textit{\textbf{Sparsity}}}  \\ \cline{2-3}  & \multicolumn{1}{c|}{0.64}  & 1.55 \\ \hline
        \end{tabular}
        \captionof{table}{\small The method-specific quality of the explanations produced for 20 randomly selected data points.}
        \label{tab:xaiquality}
    \end{minipage}%
    \hspace{0.02\textwidth}
    \begin{minipage}{0.52\textwidth}
        \centering
        \includegraphics[width=\textwidth]{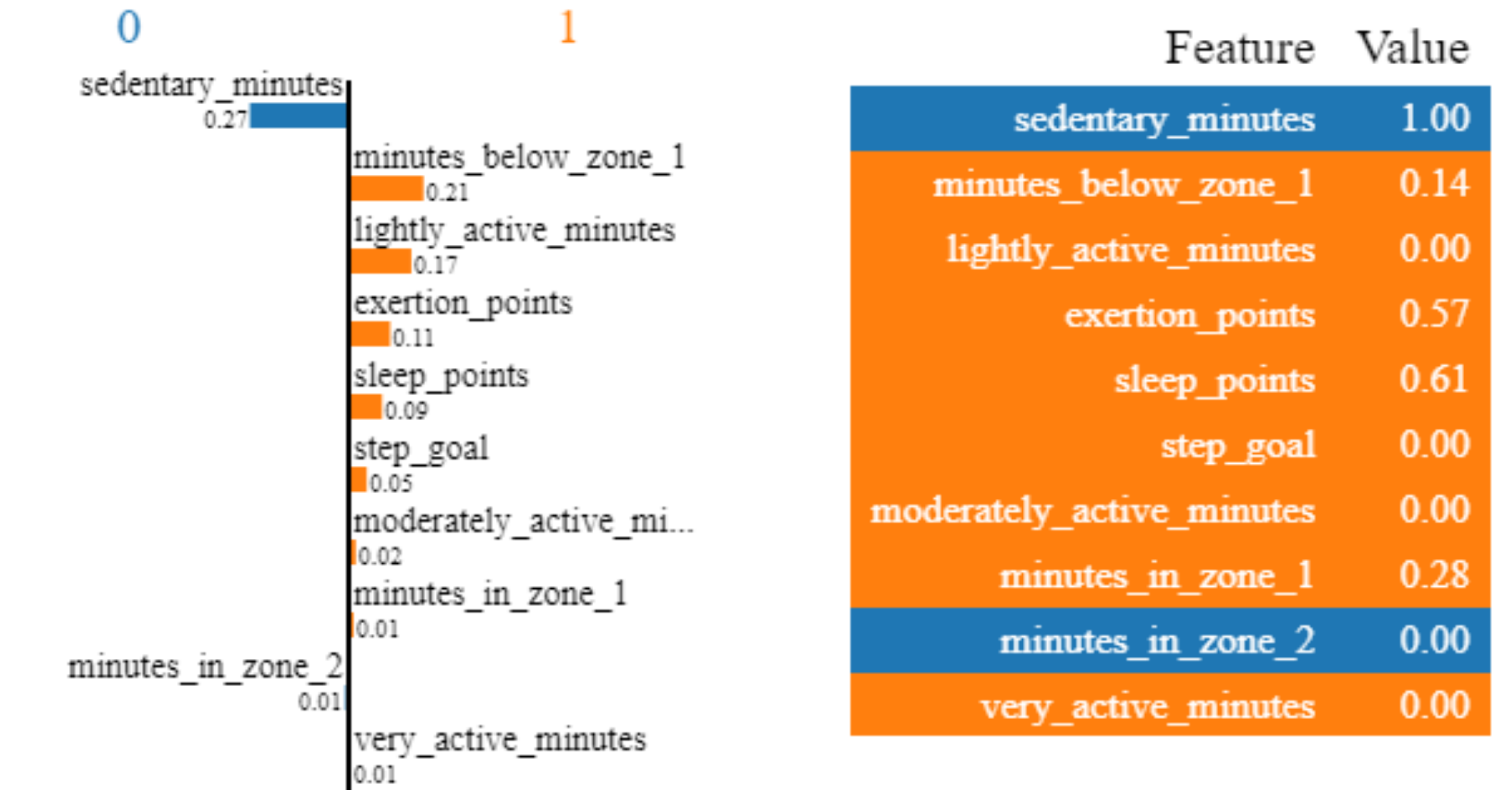}
        \caption{\small The visualization of the feature-importance-based LIME explanation of a randomly selected data point.}
        \label{fig:lime}
    \end{minipage}
\end{figure}

At first, to explain the results of the clustering analysis using XAI methods, we reformulate the clustering task as a classification problem (achieving 99\% accuracy and F1 using the Support Vector Classifier), following previous studies that use cluster labels as class targets \cite{loetsch2021interpretation,bobek2022enhancing}, and then we apply XAI methods to the classification results to generate explanations for the cluster assignments. Tab. \ref{tab:xaiquality} summarizes the average explanation quality for each XAI method, based on a sample of 20 randomly selected data points. For local methods (i.e. LIME, Anchor, Counterfactuals), quality is assessed at the instance level and averaged across the 20 data points, while for the global method (i.e. Coefficients), quality is evaluated globally, with the same score applied to all data points. We acknowledge the very high average 0.93 fidelity of LIME explanations and, thus, use them as the ground-truth explainability-relevant knowledge for our ``thesaurus''. Fig. \ref{fig:lime} indicatively presents an explanation produced by LIME, where it is evident that effectively identifying discriminatory features from this visualization is challenging, often requiring additional processing (\textbf{\textit{G3}}). Practically, most non-expert users might struggle to easily recognize that an individual's sedentary minutes negatively impact well-being, whereas minutes in zones below zone 1 and lightly active minutes have a positive effect. To summarize, all this ground-truth contextual information serves our benchmark-base which can be transferred to the LLMs via in-context learning to ensure responses that are not only domain-relevant but also grounded in explainability principles. Thus, in section \ref{methodology} we propose a systematic framework that leverages this ``thesaurus'' aiming to provide a HCXAI solution for democratizing XAI.


\section{The proposed LLM-based framework}\label{methodology}
\begin{figure}[htp]
    \centering
    \includegraphics[width=\textwidth]{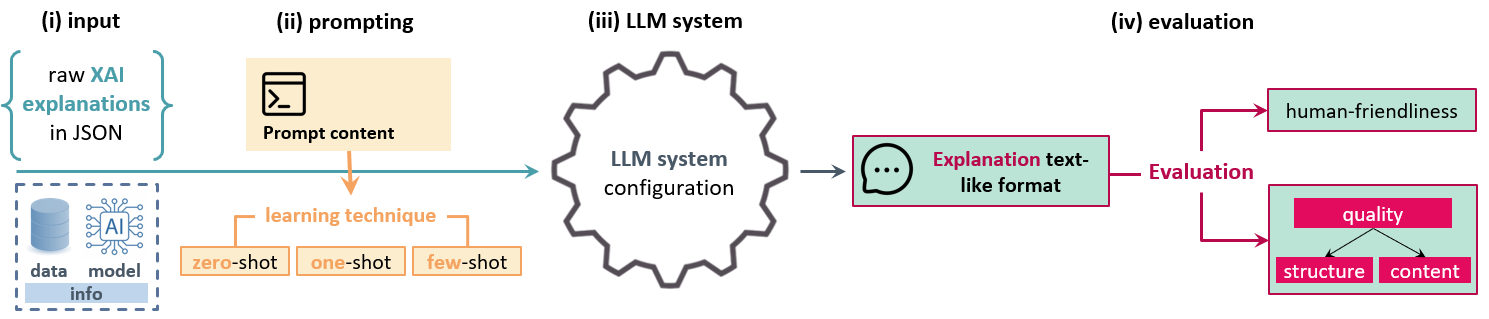}
    \caption{The high-level architecture of our framework.}
    \label{fig:llm_pipeline}
\end{figure}

We propose an in-context LLM-based framework for democratizing XAI by providing both transparency-relevant explanations to experts and human-centered explanations to non-experts in a single response (\textbf{C2}). The proposed framework is reproducible and generalizable (data-, model-, and explanations-agnostic), with its high-level architecture is presented in Fig. \ref{fig:llm_pipeline}, where: (i) receives as input the ground-truth contextual ``thesaurus'' (described in section \ref{thesaurus}) containing domain- and explainability-relevant information; (ii) uses prompting techniques (i.e. zero-shot, one-shot, few-shot) to structure and transfer the ``thesaurus'' knowledge to the LLM; (iii) configures the LLM (i.e. Mistral 7B, LLaMA3 8B) system to provide, in a single response, both transparency-related explanations for experts and human-centered explanations for non-experts; and (iv) evaluates the produced explanations for structure, content quality, and human-friendliness.

\begin{figure}[ht!]
    \centering
    \subfloat[prompt]{\label{sfig:prompt}\includegraphics[width=.499\textwidth]{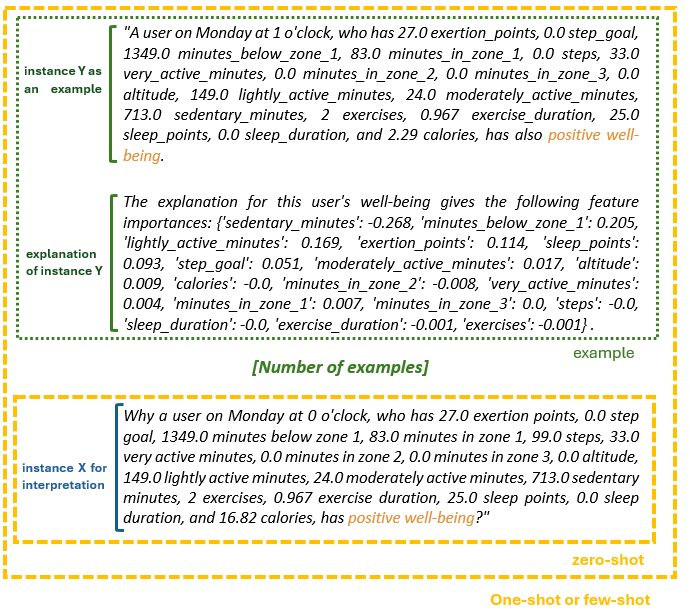}}\hfill
    \subfloat[system]{\label{sfig:system}\includegraphics[width=.499\textwidth]{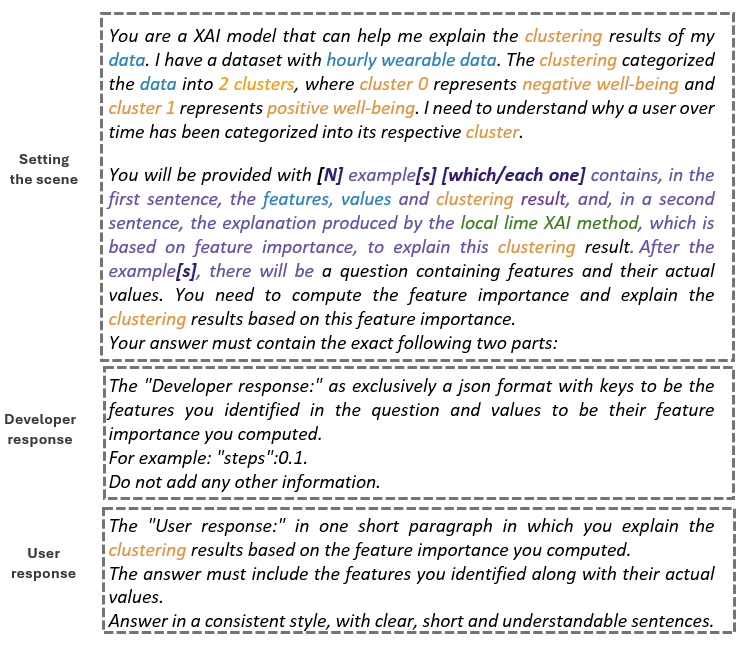}}\\
    \caption{The generalizable and configurable templates of the prompt (\ref{sfig:prompt}) and system (\ref{sfig:system}) design and setup.}
    \label{fig:promptssystem}
\end{figure}

Our ``thesaurus'' provides rich domain- (i.e. Lifesnaps multi-modal data, well-being clusters) and explainability-relevant (e.g., raw LIME explanations) knowledge, which will be transferred to the LLM using in-context learning prompting techniques, serving as a benchmark-base for our well-being monitoring in the critical healthcare domain. However, the proposed framework is generalizable due to its data-, model-, and explanation-agnostic prompt and system design. Firstly, its data-agnostic design can process data of varying granularities (e.g., hourly or daily), features (information available in the dataset), and instances (data points selected for interpretation). Secondly, with its model-agnostic design it can adapt to multiple ML tasks (e.g., clustering, classification) and targets (e.g., well-being monitoring). Thirdly, its (under-conditions) explanation-agnostic design processes explanations generated by various feature-importance-based XAI methods across different scopes (e.g., global or local). Fig. \ref{fig:promptssystem} presents the agnostic design and setup of the prompt and system templates on which our framework is built, with highlighted parts indicating configurations: light blue for data, orange for models, and light green for explanations.

Starting with the prompt template (Fig. \ref{sfig:prompt}), it is generalizable and configurable to adapt to all three prompting techniques (i.e. zero-shot, one-shot, few-shot) supported by our framework. In the zero-shot prompting setup (bottom yellow box), the LLM receives the ML model's output (i.e. well-being cluster) along with the feature values and is expected to generate an explanation without prior demonstration examples, relying heavily on the LLM's pre-trained knowledge. In contrast, the one-shot prompting setup (the outer yellow box with one iteration of the inner green box) includes the model's output, the feature values, and one demonstration example of output-feature-explanation, guiding the LLM to understand the format and context, thus providing a more coherent and relevant explanation. Finally, the few-shot prompting setup (the outer yellow box with ``few'' iterations of the inner green box) extends the one-shot prompting technique by providing multiple demonstration examples, offering a richer context for the LLM.

Moving to the system configuration template (Fig. \ref{sfig:system}), it follows the same agnostic design and prioritizes generating both transparency-related explanations for experts and human-centered explanations for non-experts in a single response. It is structured into three main parts, each serving a distinct purpose. Firstly, the ``Setting the Scene'' part introduces to the LLM its role as an explainer and provides it with summarized data-, model-, and explainability-relevant information, supporting the three different learning techniques: in zero-shot prompting, information related to explainability and demonstration examples are omitted (purple); in one-shot prompting, contextual information from the demonstration example is provided (light purple); and in few-shot prompting, additional details on the number of demonstration examples are included (dark purple). Most importantly, this part establishes a structured prompt-response mechanism, guiding the LLM in processing the data and XAI explanations, included in the prompt, to compute feature importance. Secondly, the ``Expert Response'' part guides the LLM to produce the transparency-related explanation for experts, outputting a JSON-formatted list of identified features and their importance values, which experts can analyze to understand the ``black-box'' model they have used (\textit{G1}). Finally, the ``Non-expert Response'' part guides the LLM to produce the human-centered explanation for non-experts, translating computed feature importances into a coherent, human-friendly paragraph that highlights the most influential features in the model’s decision, thus democratizing the popular LIME XAI method (\textit{G2}).

\begin{figure}[htp]
    \centering
    \includegraphics[width=0.7\textwidth]{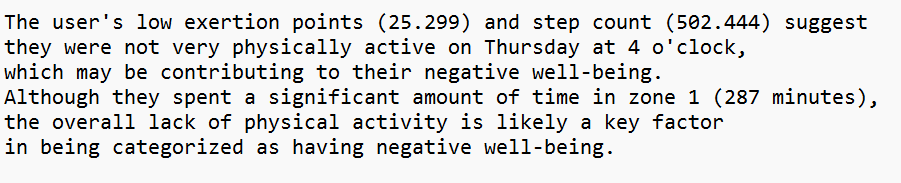}
    \caption{An indicative output human-centered explanation of our framework.}
    \label{fig:llm_response}
\end{figure}

Last but not least, thanks to our framework’s ability to generate both transparency -related and human-centered explanations in a single response, it overcomes the limitations of additional post-interpretations and lengthy sequential processes (\textit{G3}). Specifically, the three-step process required for a new instance -model inference, explanation extraction, and subsequent post-processing (see section \ref{introduction})— is now streamlined into a one-step process involving only model inference and only a few demonstration examples, facilitated by in-context learning that allows contextual adaptation even with limited demonstrations. Fig. \ref{fig:llm_response} shows the human-centered explanation generated by our framework, while Fig. \ref{fig:lime} presents the corresponding LIME explanation for the same data point. We believe the improvement in interpretability and human-friendliness for non-experts is evident, and that our framework can serve as a stepping stone toward delivering more effective HCXAI solutions —an outcome further supported by the results of the comparative human-friendliness evaluation in section \ref{evaluation}.

\section{Framework validation}\label{evaluation}
We evaluate our proposed framework by assessing the content quality of transparency -related explanations and the structure quality of human-centered explanations (section \ref{quality}), as well as the human-friendliness of the human-centered explanations (section \ref{humanfriendliness}) (\textbf{C3}).

\subsection{Human-friendliness evaluation} \label{humanfriendliness}
To evaluate the human-friendliness of the human-centered explanations produced by our proposed framework, we conducted a user study (N=56) in which participants compared our explanations (generated using few-shot prompting with LLaMA3 8B) to those produced by the LIME XAI method. To do so, we used the short version of the User Experience Questionnaire (UEQ) \cite{laugwitz2008construction}, which includes two primary scales: \textit{Pragmatic Quality (PQ)}, including items related to usefulness, efficiency, and task orientation; and \textit{Hedonic Quality (HQ)}, with items related to enjoyment, engagement, and stimulation. Each item is rated on a 7-point semantic differential scale between two opposing adjectives, ranging from -3 (very negative evaluation) to 3 (very positive evaluation). 

The 56 participants in our study were randomly recruited via the authors' university mailing list and professional and social media networks. The majority (39) were from STEM fields, while 7 were from Health \& Sport Sciences, 6 from Social Sciences \& Humanities, and 4 from Applied Sciences \& Engineering. We note that no personal information was collected during the study and all responses were anonymous and securely stored on the university's infrastructure. Prior to participation, participants were informed about the study's purpose, introduced to key concepts such as well-being clustering analysis, LLMs, and LIME, and provided their consent. The first part of the study was an A/B test, where participants were randomly assigned to either group A or group B with those in group A receiving explanations generated by LIME (Fig. \ref{fig:lime}), while those in group B receiving explanations generated by our framework (Fig. \ref{fig:llm_response}), but in both cases, participants rated their experience using the UEQ short. The second part of the study was a paired comparison, where all participants, regardless of their initial group assignment, interacted with both LIME and our explanations and re-evaluated their experience using the UEQ short. This mixed approach —combining A/B testing and paired comparisons— ensures unbiased responses and enables direct comparisons for deeper insights.

\begin{table}[ht!]
\centering
\tiny
\caption{The UEQ results of the statistical analysis of the A/B testing for comparing the user experience of the LIME explanations (feedback from the 25 participants in group A) and our framework explanations (feedback from the 31 participants in group B). Values can range between -3 and 3 and best mean values are in bold.}
\label{tab:AB_UEQ}
\begin{tabular}{cc|ccccc||ccccc}
\hline
\multicolumn{1}{l}{} &
  \multicolumn{1}{l|}{} &
  \multicolumn{5}{c||}{\textit{\textbf{LIME explanations}}} &
  \multicolumn{5}{c}{\textit{\textbf{Our framework explanations}}} \\ \hline
\multicolumn{1}{c|}{\textbf{Scale}} &
  \textbf{Item} &
  \multicolumn{1}{c|}{\textbf{Mean}} &
  \multicolumn{1}{c|}{\textbf{Std. Dev.}} &
  \multicolumn{1}{c|}{\textbf{Conf.}} &
  \multicolumn{2}{c||}{\textbf{Conf. Int.}} &
  \multicolumn{1}{c|}{\textbf{Mean}} &
  \multicolumn{1}{c|}{\textbf{Std. Dev.}} &
  \multicolumn{1}{c|}{\textbf{Conf.}} &
  \multicolumn{2}{c}{\textbf{Conf. Int.}} \\ \hline \hline 
  \multicolumn{1}{c|}{\multirow{4}{*}{\textit{\begin{tabular}[c]{@{}c@{}}Pragmatic\\ Quality\end{tabular}}}} &
  obstructive-supportive &
  \multicolumn{1}{c|}{0.32} &
  \multicolumn{1}{c|}{1.57} &
  \multicolumn{1}{c|}{0.61} &
  \multicolumn{1}{c|}{-0.29} &
   0.93 &
  \multicolumn{1}{c|}{\textbf{0.35}} &
  \multicolumn{1}{c|}{1.47} &
  \multicolumn{1}{c|}{0.51} &
  \multicolumn{1}{c|}{-0.16} &
  0.87 \\ \cline{2-12} 
\multicolumn{1}{c|}{} &
  complicated-easy &
  \multicolumn{1}{c|}{-0.08} &
  \multicolumn{1}{c|}{1.84} &
  \multicolumn{1}{c|}{0.72} &
  \multicolumn{1}{c|}{-0.80} &
  0.64 &
  \multicolumn{1}{c|}{\textbf{0.29}} &
  \multicolumn{1}{c|}{1.84} &
  \multicolumn{1}{c|}{0.65} &
  \multicolumn{1}{c|}{-0.36} &
  0.94 \\ \cline{2-12} 
\multicolumn{1}{c|}{} &
  inefficient-efficient &
  \multicolumn{1}{c|}{0.32} &
  \multicolumn{1}{c|}{1.57} &
  \multicolumn{1}{c|}{0.61} &
  \multicolumn{1}{c|}{-0.29} &
  0.93 &
  \multicolumn{1}{c|}{\textbf{0.48}} &
  \multicolumn{1}{c|}{1.82} &
  \multicolumn{1}{c|}{0.64} &
  \multicolumn{1}{c|}{-0.15} &
  1.12 \\ \cline{2-12} 
\multicolumn{1}{c|}{} &
  confusing-clear &
  \multicolumn{1}{c|}{0.12} &
  \multicolumn{1}{c|}{1.94} &
  \multicolumn{1}{c|}{0.76} &
  \multicolumn{1}{c|}{-0.64} &
  0.88 &
  \multicolumn{1}{c|}{\textbf{0.71}} &
  \multicolumn{1}{c|}{1.84} &
  \multicolumn{1}{c|}{0.65} &
  \multicolumn{1}{c|}{0.05} &
  1.36 \\ \hline
\multicolumn{2}{r|}{\textit{\textbf{Pragmatic Quality Overall}}} &
  \multicolumn{1}{c|}{0.17} &
  \multicolumn{1}{c|}{1.39} &
  \multicolumn{1}{c|}{0.54} &
  \multicolumn{1}{c|}{-0.37} &
  0.71 &
  \multicolumn{1}{c|}{\textbf{0.46}} &
  \multicolumn{1}{c|}{1.51} &
  \multicolumn{1}{c|}{0.53} &
  \multicolumn{1}{c|}{-0.07} &
  0.99 \\ \hline \hline
\multicolumn{1}{c|}{\multirow{4}{*}{\textit{\begin{tabular}[c]{@{}c@{}}Hedonic \\ Quality\end{tabular}}}} &
  boring-exciting &
  \multicolumn{1}{c|}{-0.56} &
  \multicolumn{1}{c|}{1.85} &
  \multicolumn{1}{c|}{0.72} &
  \multicolumn{1}{c|}{-1.28} &
  0.16 &
  \multicolumn{1}{c|}{\textbf{-0.35}} &
  \multicolumn{1}{c|}{1.38} &
  \multicolumn{1}{c|}{0.48} &
  \multicolumn{1}{c|}{-0.84} &
  0.13 \\ \cline{2-12} 
\multicolumn{1}{c|}{} &
  not interesting-interesting &
  \multicolumn{1}{c|}{-0.12} &
  \multicolumn{1}{c|}{1.98} &
  \multicolumn{1}{c|}{0.77} &
  \multicolumn{1}{c|}{-0.89} &
  0.65 &
  \multicolumn{1}{c|}{\textbf{0.09}} &
  \multicolumn{1}{c|}{1.59} &
  \multicolumn{1}{c|}{0.56} &
  \multicolumn{1}{c|}{-0.46} &
  0.66 \\ \cline{2-12} 
\multicolumn{1}{c|}{} &
  conventional-inventive &
  \multicolumn{1}{c|}{\textbf{-0.40}} &
  \multicolumn{1}{c|}{1.50} &
  \multicolumn{1}{c|}{0.58} &
  \multicolumn{1}{c|}{-0.98} &
  0.18 &
  \multicolumn{1}{c|}{-0.48} &
  \multicolumn{1}{c|}{1.45} &
  \multicolumn{1}{c|}{0.51} &
  \multicolumn{1}{c|}{-0.99} &
  0.02 \\ \cline{2-12} 
\multicolumn{1}{c|}{} &
  usual-leading edge &
  \multicolumn{1}{c|}{\textbf{-0.32}} &
  \multicolumn{1}{c|}{1.46} &
  \multicolumn{1}{c|}{0.57} &
  \multicolumn{1}{c|}{-0.89} &
  0.25 &
  \multicolumn{1}{c|}{-0.54} &
  \multicolumn{1}{c|}{1.31} &
  \multicolumn{1}{c|}{0.46} &
  \multicolumn{1}{c|}{-1.01} &
  -0.08 \\ \hline
\multicolumn{2}{r|}{\textit{\textbf{Hedonic Quality Overall}}} &
  \multicolumn{1}{c|}{-0.35} &
  \multicolumn{1}{c|}{1.44} &
  \multicolumn{1}{c|}{0.56} &
  \multicolumn{1}{c|}{-0.91} &
  0.21 &
  \multicolumn{1}{c|}{\textbf{-0.32}} &
  \multicolumn{1}{c|}{1.04} &
  \multicolumn{1}{c|}{0.36} &
  \multicolumn{1}{c|}{-0.69} &
  0.04 \\ \hline
\end{tabular}
\end{table}

Tab. \ref{tab:AB_UEQ} presents the A/B testing results, where —despite the absence of statistical significant differences (see Conf. Int.)— a trend emerges favoring our explanations, as reflected in higher mean values and overall scores for both PQ (0.46 vs. 0.17) and HQ (-0.32 vs. -0.35). More specifically, our framework scores higher across all PQ items, including supportiveness (0.35 vs. 0.32), ease of use (0.29 vs. -0.08), efficiency (0.48 vs. 0.32), and clarity (0.71 vs. 0.12), indicating modest but positive trends in user preference. While our explanations also show a slight advantage in HQ, driven by higher scores in excitement (-0.35 vs. -0.56) and interest (0.09 vs. -0.12), the overall HQ difference remains marginal (-0.32 vs. -0.35), mainly due to lower perceived inventiveness (-0.48 vs. -0.40) and lower scores for being seen as leading-edge (-0.54 vs. -0.32), and the overall HQ remains relatively limited.

\begin{table}[ht!]
\centering
\tiny
\caption{The UEQ results of the statistical analysis of the paired comparison between the user experience of the LIME explanations and our framework explanations evaluated side-by-side. Values can range between -3 and 3 and best mean values are in bold.}
\label{tab:paired_UEQ}
\begin{tabular}{cc|ccccc||ccccc}
\hline
\multicolumn{1}{l}{} &
  \multicolumn{1}{l|}{} &
  \multicolumn{5}{c||}{\textit{\textbf{LIME explanations}}} &
  \multicolumn{5}{c}{\textit{\textbf{Our framework explanations}}} \\ \hline
\multicolumn{1}{c|}{\textbf{Scale}} &
  \textbf{Item} &
  \multicolumn{1}{c|}{\textbf{Mean}} &
  \multicolumn{1}{c|}{\textbf{Std. Dev.}} &
  \multicolumn{1}{c|}{\textbf{Conf.}} &
  \multicolumn{2}{c||}{\textbf{Conf. Int.}} &
  \multicolumn{1}{c|}{\textbf{Mean}} &
  \multicolumn{1}{c|}{\textbf{Std. Dev.}} &
  \multicolumn{1}{c|}{\textbf{Conf.}} &
  \multicolumn{2}{c}{\textbf{Conf. Int.}} \\ \hline \hline 
  \multicolumn{1}{c|}{\multirow{4}{*}{\textit{\begin{tabular}[c]{@{}c@{}}Pragmatic\\ Quality\end{tabular}}}} &
  obstructive-supportive &
  \multicolumn{1}{c|}{-0.14} &
  \multicolumn{1}{c|}{1.56} &
  \multicolumn{1}{c|}{0.41} &
  \multicolumn{1}{c|}{-0.55} &
   0.26 &
  \multicolumn{1}{c|}{\textbf{0.92}} &
  \multicolumn{1}{c|}{1.42} &
  \multicolumn{1}{c|}{0.37} &
  \multicolumn{1}{c|}{0.55} &
  1.30 \\ \cline{2-12} 
\multicolumn{1}{c|}{} &
  complicated-easy &
  \multicolumn{1}{c|}{-0.28} &
  \multicolumn{1}{c|}{1.94} &
  \multicolumn{1}{c|}{0.50} &
  \multicolumn{1}{c|}{-0.79} &
  0.22 &
  \multicolumn{1}{c|}{\textbf{1.10}} &
  \multicolumn{1}{c|}{1.51} &
  \multicolumn{1}{c|}{0.39} &
  \multicolumn{1}{c|}{0.71} &
  1.50 \\ \cline{2-12} 
\multicolumn{1}{c|}{} &
  inefficient-efficient &
  \multicolumn{1}{c|}{0.08} &
  \multicolumn{1}{c|}{1.59} &
  \multicolumn{1}{c|}{0.41} &
  \multicolumn{1}{c|}{-0.32} &
  0.50 &
  \multicolumn{1}{c|}{\textbf{0.94}} &
  \multicolumn{1}{c|}{1.32} &
  \multicolumn{1}{c|}{0.34} &
  \multicolumn{1}{c|}{0.59} &
  1.29 \\ \cline{2-12} 
\multicolumn{1}{c|}{} &
  confusing-clear &
  \multicolumn{1}{c|}{-0.30} &
  \multicolumn{1}{c|}{1.70} &
  \multicolumn{1}{c|}{0.44} &
  \multicolumn{1}{c|}{-0.75} &
  0.14 &
  \multicolumn{1}{c|}{\textbf{1.05}} &
  \multicolumn{1}{c|}{1.62} &
  \multicolumn{1}{c|}{0.42} &
  \multicolumn{1}{c|}{0.62} &
  1.47 \\ \hline
\multicolumn{2}{r|}{\textit{\textbf{Pragmatic Quality Overall}}} &
  \multicolumn{1}{c|}{-0.16} &
  \multicolumn{1}{c|}{1.52} &
  \multicolumn{1}{c|}{0.39} &
  \multicolumn{1}{c|}{-0.55} &
  0.23 &
  \multicolumn{1}{c|}{\textbf{1.00}} &
  \multicolumn{1}{c|}{1.25} &
  \multicolumn{1}{c|}{0.33} &
  \multicolumn{1}{c|}{0.67} &
  1.33 \\ \hline \hline
\multicolumn{1}{c|}{\multirow{4}{*}{\textit{\begin{tabular}[c]{@{}c@{}}Hedonic \\ Quality\end{tabular}}}} &
  boring-exciting &
  \multicolumn{1}{c|}{-0.46} &
  \multicolumn{1}{c|}{1.67} &
  \multicolumn{1}{c|}{0.43} &
  \multicolumn{1}{c|}{-0.90} &
  -0.02 &
  \multicolumn{1}{c|}{\textbf{-0.12}} &
  \multicolumn{1}{c|}{1.29} &
  \multicolumn{1}{c|}{0.33} &
  \multicolumn{1}{c|}{-0.46} &
  0.21 \\ \cline{2-12} 
\multicolumn{1}{c|}{} &
  not interesting-interesting &
  \multicolumn{1}{c|}{-0.07} &
  \multicolumn{1}{c|}{1.74} &
  \multicolumn{1}{c|}{0.45} &
  \multicolumn{1}{c|}{-0.52} &
  0.38 &
  \multicolumn{1}{c|}{\textbf{0.42}} &
  \multicolumn{1}{c|}{1.47} &
  \multicolumn{1}{c|}{0.38} &
  \multicolumn{1}{c|}{0.04} &
  0.81 \\ \cline{2-12} 
\multicolumn{1}{c|}{} &
  conventional-inventive &
  \multicolumn{1}{c|}{-0.17} &
  \multicolumn{1}{c|}{1.50} &
  \multicolumn{1}{c|}{0.39} &
  \multicolumn{1}{c|}{-0.57} &
  0.21 &
  \multicolumn{1}{c|}{\textbf{0.14}} &
  \multicolumn{1}{c|}{1.38} &
  \multicolumn{1}{c|}{0.36} &
  \multicolumn{1}{c|}{-0.21} &
  0.50 \\ \cline{2-12} 
\multicolumn{1}{c|}{} &
  usual-leading edge &
  \multicolumn{1}{c|}{-0.37} &
  \multicolumn{1}{c|}{1.35} &
  \multicolumn{1}{c|}{0.35} &
  \multicolumn{1}{c|}{-0.73} &
  -0.02 &
  \multicolumn{1}{c|}{\textbf{0}} &
  \multicolumn{1}{c|}{1.38} &
  \multicolumn{1}{c|}{0.36} &
  \multicolumn{1}{c|}{-0.36} &
  0.36 \\ \hline
\multicolumn{2}{r|}{\textit{\textbf{Hedonic Quality Overall}}} &
  \multicolumn{1}{c|}{-0.27} &
  \multicolumn{1}{c|}{1.36} &
  \multicolumn{1}{c|}{0.35} &
  \multicolumn{1}{c|}{-0.63} &
  0.08 &
  \multicolumn{1}{c|}{\textbf{0.11}} &
  \multicolumn{1}{c|}{1.13} &
  \multicolumn{1}{c|}{0.29} &
  \multicolumn{1}{c|}{-0.18} &
  0.40 \\ \hline
\end{tabular}
\end{table}

Tab. \ref{tab:paired_UEQ} presents the results of the paired analysis, revealing a strong preference for our framework’s explanations over LIME’s (see Conf. Int.). This statistical significant improvement in perceived user experience applies not only to the overall PQ score (1.00 vs. -0.16) but also to all individual PQ items, including supportiveness (0.92 vs. -0.14), ease of use (1.10 vs. -0.28), efficiency (0.94 vs. 0.08), and clarity (1.05 vs. -0.30). While HQ shows a slight improvement (0.11 vs. -0.27), this difference is not statistical significant, and the overall user experience remains relatively limited. However, our explanations receive higher scores across all HQ items, including excitement (-0.12 vs. -0.46), interest (0.42 vs. -0.07), perceived inventiveness (0.14 vs. -0.17), and being leading-edge (0 vs. -0.37).

We also included open-ended questions, but we did not collect enough responses to conduct a thorough qualitative analysis. Some key manual observations revealed that LIME’s visualizations were difficult for non-experts to interpret, whereas our framework’s explanations were generally more accessible and easier to understand. Nevertheless, participants suggested that these explanations could be improved with additional context and visual elements. Overall, it is worth emphasizing the statistical significant improvement in the overall PQ score (1.00 vs. -0.16) and across all its items, showcasing our framework’s ability to deliver human-centered, understandable explanations for non-experts.

\subsection{Quality evaluation}\label{quality}
The above user study focused on evaluating the human-friendliness of our framework’s human-centered explanations and to complement this, we additionally assess their structure quality and the content quality of the transparency-related explanations. Drawing on established literature, we select appropriate metrics to evaluate the quality of LLM-generated responses, which reflect the effectiveness of the explanations produced by our framework. First, the \textit{structure quality} evaluation examines how well the responses (i.e. explanations) adhere to the requested information (i.e. prompt), ensuring consistency and linguistic appropriateness, and thus, we employ the following metrics \cite{vadlapudi2010automated,pitler2010automatic}: \textit{coherence} \cite{fang2016using}, which measures content similarity between the prompt and the explanation, ranging from -1 (opposite) to 1 (identical); \textit{grammar errors}\footnote{\url{https://github.com/jxmorris12/language\_tool\_python}}, which counts the number of grammatical mistakes in the explanation; \textit{Automated Readability Index (ARI)} \cite{senter1967automated}, which estimates the difficulty level of the explanation; and \textit{sentiment consistency}\footnote{\url{https://textblob.readthedocs.io/en/dev/}}, which assesses how the emotional tone of the explanation aligns with that of the prompt, ranging from 0 (identical) to 2 (maximum difference). The \textit{content quality} evaluation quantifies the similarity between explanations generated by the ground-truth XAI method (i.e., LIME) and those generated by our framework, and thus, we use the following metrics:\textit{ Spearman rank correlation} \cite{sedgwick2014spearman}, which measures the association between the two (LIME vs. our framework) ranked feature importance sets, ranging from -1 (perfect inverse) to 1 (perfect agreement); \textit{Normalized Discounted Cumulative Gain (NDCG) difference} \cite{jarvelin2002cumulated}, which measures the quality of feature relevance and ranking, ranging from 0 (identical) to 1 (totally dissimilar); and \textit{Euclidean distance}, which calculates the distance between the two feature importance vectors. Tab. \ref{tab:llmquality} summarizes the average structure and content quality of the explanations produced by our framework for 20 randomly selected instances, both for Mistral and LLaMA3, across the three prompting techniques, zero-, one-, and few-shot prompting. 

\begin{table}[htp]
\centering
\scriptsize
\caption{The average structure and content quality of 20 randomly selected instances across the three learning techniques and two models.}
\label{tab:llmquality}
\begin{tabular}{r|c|cccc|ccc}
\hline
\multicolumn{1}{l|}{} &
  \multicolumn{1}{l|}{} &
  \multicolumn{4}{c|}{\textbf{structure}} &
  \multicolumn{3}{c}{\textbf{Content}} \\ \hline
\multicolumn{1}{c|}{\textbf{Technique}} &
  \textbf{LLM} &
  \multicolumn{1}{c|}{\textit{coherence}} &
  \multicolumn{1}{c|}{\textit{\begin{tabular}[c]{@{}c@{}}grammar\\ errors\end{tabular}}} &
  \multicolumn{1}{c|}{\textit{readability}} &
  \multicolumn{1}{c|}{\textit{\begin{tabular}[c]{@{}c@{}}sentiment\\ consistency\end{tabular}}} &
  \multicolumn{1}{c|}{\textit{\begin{tabular}[c]{@{}c@{}}spearman\\ rank corr.\end{tabular}}} &
  \multicolumn{1}{c|}{\textit{\begin{tabular}[c]{@{}c@{}}NDCG\\ difference\end{tabular}}} &
  \textit{\begin{tabular}[c]{@{}c@{}}Euclidean\\ distance\end{tabular}} \\ \hline \hline
\multirow{2}{*}{zero-shot} &
  Mistral & \multicolumn{1}{c|}{0.71} & \multicolumn{1}{c|}{0.15} & \multicolumn{1}{c|}{45.45} &
  \multicolumn{1}{c|}{0.15} & \multicolumn{1}{c|}{0.13} & \multicolumn{1}{c|}{0.08} & 1.02 \\ \cline{2-9} &
  LLaMA3 & \multicolumn{1}{c|}{\textbf{0.77}} & \multicolumn{1}{c|}{\textbf{0}} & \multicolumn{1}{c|}{44.55} &
  \multicolumn{1}{c|}{0.17} & \multicolumn{1}{c|}{0.01} & \multicolumn{1}{c|}{0.07} & 0.32 \\ \hline
\multirow{2}{*}{one-shot} &
  Mistral & \multicolumn{1}{c|}{0.72} & \multicolumn{1}{c|}{0.77} & \multicolumn{1}{c|}{\textbf{40.58}} &
  \multicolumn{1}{c|}{0.20} & \multicolumn{1}{c|}{0.44} & \multicolumn{1}{c|}{0.02} &  0.34 \\ \cline{2-9}  &
  LLaMA3 & \multicolumn{1}{c|}{0.76} & \multicolumn{1}{c|}{0.21} & \multicolumn{1}{c|}{45.95} &
  \multicolumn{1}{c|}{0.18} & \multicolumn{1}{c|}{0.91} & \multicolumn{1}{c|}{0.009} &  0.05 \\ \hline
\multirow{2}{*}{few-shot} &
Mistral & \multicolumn{1}{c|}{0.71} & \multicolumn{1}{c|}{0.73} & \multicolumn{1}{c|}{50.38} &
  \multicolumn{1}{c|}{0.20} & \multicolumn{1}{c|}{0.74} & \multicolumn{1}{c|}{0.012} & 0.20 \\ \cline{2-9}  &
  LLaMA3 & \multicolumn{1}{c|}{0.71} & \multicolumn{1}{c|}{0.81} & \multicolumn{1}{c|}{43.64} &
  \multicolumn{1}{c|}{\textbf{0.27}} & \multicolumn{1}{c|}{\textbf{0.92}} & \multicolumn{1}{c|}{\textbf{0.001}} &  \textbf{0.02} \\ \hline
\end{tabular}
\end{table}

The \textit{structure quality} of our human-centered explanations shows mixed results, varying by LLM and prompting technique. In terms of coherence, Mistral shows a slight improvement from 0.71 in zero-shot to 0.72 in one-shot, but drops back to 0.71 in few-shot. LLaMA3's coherence, on the other hand, decreases from 0.77 in zero-shot to 0.71 in few-shot. Grammatical errors increase for LLaMA3, reaching 0.81 in the few-shot scenario, while for Mistral, the errors initially rise but later decrease slightly. Readability shows varying trends, with Mistral’s readability declining in the few-shot setting, making the text harder to follow, while LLaMA3 improves readability in the same scenario. The only metric with a consistently positive trend is sentiment consistency. Mistral shows an increase from 0.15 in zero-shot to 0.20 in one-shot, remaining stable in few-shot, while LLaMA3 improves from 0.17 in zero-shot to 0.18 in one-shot, then jumps to 0.27 in few-shot. These variations suggest that the structure quality is influenced more by the intrinsic capabilities of each LLM rather than the specific prompting techniques or the amount of contextual information provided. 

Regarding \textit{content quality}, both LLMs show significant improvements when provided with more contextual information. For instance, LLaMA3's Spearman rank correlation improves dramatically from 0.01 in zero-shot to 0.92 in few-shot, while Mistral's correlation increases from 0.13 to 0.74. The NDCG score (where lower values indicate better relevance) improves for both models as they transition from zero-shot to few-shot, with LLaMA3 improving from 0.07 to 0.001, and Mistral from 0.08 to 0.012. Additionally, Euclidean distance (where lower values reflect greater similarity) decreases significantly for both models as they move from zero-shot to few-shot. These findings underscore that increasing context through few-shot techniques significantly enhances the content quality, improving alignment, relevance, and accuracy, consistent with previous research \cite{cahyawijaya2024llms,liu2023large}. 

In comparing Mistral and LLaMA3, each model demonstrates distinct strengths and weaknesses. Mistral remains more stable in coherence and grammatical accuracy across different prompting techniques, whereas LLaMA3 excels in readability and content quality, particularly in the few-shot setting. LLaMA3 also shows a closer alignment with LIME's ground-truth explanations, further highlighting its robustness and effectiveness as a model-agnostic explainer when provided with additional context. Therefore, our results suggest that few-shot prompting significantly enhances the overall quality of explanations.

\section{Discussion \& Limitations}\label{discussion}
Complementing the evaluations presented in section \ref{evaluation}, we explore critical issues regarding the use of LLMs, assessing whether these are present in our framework. Specifically, we examine: (i) whether LLM responses exhibit hallucinations; (ii) the black-box nature of LLMs in computing feature importance; and (iii) the consistency within LLM responses. This discussion does not aim to offer a comprehensive assessment but rather to highlight the significance of these issues and emphasize the need for further research toward a more thorough evaluation.

\textbf{Do our explanations exhibit hallucinations?} 
The issue of hallucinations in LLM responses, where models generate information not present in the original prompt, resulting in inaccuracies, is a crucial concern in their application \cite{fang2024large,yang2024harnessing}. We evaluate hallucinations by advising the content quality of LLM explanations, especially Spearman rank correlation, and NDCG difference metrics. The results in Table \ref{tab:llmquality} indicate that advanced prompting techniques, such as few-shot prompting, can reduce hallucinations by providing additional context, thereby producing more accurate and grounded explanations, in line with conventional XAI methods. More detailed prompts improve the alignment between our framework's explanations feature importance rankings and those from the LIME method, as evidenced by greater quality in both metrics. However, hallucinations may not be entirely eliminated, highlighting the need for further research to refine LLM prompting strategies for consistently reliable outputs. 

\textbf{What is hidden behind our explanations?}
A critical point is understanding LLMs transparency, particularly regarding how they compute feature importance for generating explanations \cite{ramlochan2024blackbox}. To do so, we compared the feature importance computed by the LLaMA3 model using zero-shot and one-shot techniques with that derived from the LIME method. For one random instance, LLaMA3's zero-shot explanation, based on \textit{correlations}, showed a significant deviation from LIME's feature ranking. However, when using one-shot prompting that incorporated LIME-based explanations, the results aligned more closely with LIME. Similar patterns were observed in a second instance, where the zero-shot explanation, based on \textit{relative influence}, again deviated from LIME, but one-shot prompting led to better alignment. These findings indicate that LLMs struggle to accurately compute feature importance without prior context, but contextual learning significantly improves their performance, emphasizing the need for further research to enhance the ability of LLMs to perform well even under zero-shot prompting. 

\textbf{Is our framework within-explanation consistent?}
Another interesting point is whether LLMs can maintain consistency within responses that consist of multiple parts (Figure~\ref{sfig:system}), specifically by providing feature importance information to developers and then explaining these features to users in an understandable format (see Figure \ref{sfig:system}). Using one-shot prompting and examining the LLaMA3 and Mistral models on randomly selected instances, we find that the features highlighted in the user response corresponded to the important features identified in the developer response. This preliminary examination suggests that while both models can maintain consistency between developer and user responses, further research is needed to understand how LLMs handle chain-like queries and sustain consistency across multiple interactions.

\textbf{Limitations}.
The questions raised in this discussion highlight key directions for future research to enhance the robustness of our framework. Additionally, the current implementation supports only realistic features and future work could explore the integration of non-meaningful features (e.g., those derived from dimensionality reduction). Moreover, we plan to extend the framework to support a broader range of XAI methods beyond numerical, feature-importance-based ones. Finally, a more comprehensive evaluation is needed to improve the framework’s generalizability, including assessing factual errors in explanations, optimizing the selection of few-shot examples to ensure target instance representativeness, and analyzing more prompting techniques and LLM configurations.

\section{Conclusion}\label{conclusion}
As AI becomes increasingly embedded in critical decision-making systems, its opaque, ``black-box'' models raise urgent concerns around transparency and the lack of understandable XAI methods for non-experts, highlighting the need for truly human-centered XAI (HCXAI) solutions. In this work, we propose a generalizable, reproducible HCXAI framework that leverages LLMs and in-context learning to deliver explanations tailored to both experts and non-experts. By integrating domain- and explanation-relevant knowledge into prompts and system configurations, our framework provides transparency-related explanations for experts and human-centered explanations for non-experts in a single response. To demonstrate its effectiveness, we apply the framework to a critical well-being monitoring scenario, establishing a ground-truth contextual ``thesaurus'' through rigorous benchmarking with over 40 data, model, and XAI combinations. Our framework achieves strong content quality, closely aligning with the LIME XAI method (Spearman rank correlation = 0.92), and significantly outperforms LIME in a user study (N=56), with notably higher scores in supportiveness, ease of use, efficiency, and clarity, particularly in the paired comparison. These results confirm the potential of LLMs as enablers of HCXAI, bridging the gap between algorithmic transparency and human-centered explainability.

\appendix

\section{The LifeSnaps dataset}\label{appendixA}
\subsection{Preprocessing \& Engineering}\label{preprocessingappendix}

In section \ref{applicationdomain}, we briefly outlined the preprocessing and feature engineering steps applied to the LifeSnaps dataset. Tab. \ref{tab:preprocessingsteps} provides a more detailed breakdown of these preprocessing steps for each feature. The first two columns (feature granularity and data aggregation) are completed for all features in the training and validation sets. Conversely, the last three columns (data type conversion, granularity processing, and handling of NaNs and zeros) are only filled for features in the training set. The symbol ``-'' indicates that a specific step was deemed unnecessary after evaluation, while the ``*'' denotes that the step was not assessed for the feature because it belongs to the validation set.

\begin{table}[htp]
\tiny
\centering
\caption{Pre-processing and engineering steps in each feature of the LifeSnaps dataset.}
\label{tab:preprocessingsteps}
\begin{tabular}{cl|lllll}
\hline
\multicolumn{2}{c|}{\textbf{}} &
  \multicolumn{5}{c}{\textbf{Pre-processing steps}} \\ \hline
\multicolumn{1}{c|}{\textbf{Category}} &
  \multicolumn{1}{c|}{\textbf{Feature}} &
  \multicolumn{1}{c|}{\textbf{\begin{tabular}[c]{@{}c@{}}feature\\ granularity\end{tabular}}} &
  \multicolumn{1}{c|}{\textbf{\begin{tabular}[c]{@{}c@{}}data\\ aggregation\end{tabular}}} &
  \multicolumn{1}{c|}{\textbf{\begin{tabular}[c]{@{}c@{}}data type\\ conversion\end{tabular}}} &
  \multicolumn{1}{l|}{\textbf{\begin{tabular}[c]{@{}c@{}}granularity\\ processing\end{tabular}}} &
  \multicolumn{1}{c}{\textbf{\begin{tabular}[c]{@{}c@{}}NaN and 0\\ handling\end{tabular}}} \\ \hline \hline
  \multicolumn{1}{c|}{\multirow{10}{*}{\begin{tabular}[c]{@{}c@{}}Physical\\ Activity\end{tabular}}} &
  steps &
  \multicolumn{1}{c|}{3-minutes} &
  \multicolumn{1}{c|}{sum} &
  \multicolumn{1}{c|}{-} &
  \multicolumn{1}{c|}{-} &
  \multicolumn{1}{c}{mean} \\ \cline{2-7} 
\multicolumn{1}{c|}{} &
  altitude &
  \multicolumn{1}{c|}{arbitrary} &
  \multicolumn{1}{c|}{sum} &
  \multicolumn{1}{c|}{-} &
  \multicolumn{1}{c|}{forward} &
  \multicolumn{1}{c}{zero} \\ \cline{2-7} 
\multicolumn{1}{c|}{} &
  distance &
  \multicolumn{1}{c|}{3-minutes} &
  \multicolumn{1}{c|}{sum} &
  \multicolumn{1}{c|}{-} &
  \multicolumn{1}{c|}{-} &
  \multicolumn{1}{c}{mean} \\ \cline{2-7} 
\multicolumn{1}{c|}{} &
  \begin{tabular}[c]{@{}l@{}}sedentary/lightly/moderately\\ /very active minutes\end{tabular} &
  \multicolumn{1}{c|}{daily} &
  \multicolumn{1}{c|}{mean} &
  \multicolumn{1}{c|}{-} &
  \multicolumn{1}{c|}{daily} &
  \multicolumn{1}{c}{mean} \\ \cline{2-7} 
\multicolumn{1}{c|}{} &
  minutes below zone 1 &
  \multicolumn{1}{c|}{2-day} &
  \multicolumn{1}{c|}{mean} &
  \multicolumn{1}{c|}{-} &
  \multicolumn{1}{c|}{daily} &
  \multicolumn{1}{c}{mean} \\ \cline{2-7} 
\multicolumn{1}{c|}{} &
  minutes in zone 1/2/3 &
  \multicolumn{1}{c|}{2-day} &
  \multicolumn{1}{c|}{mean} &
  \multicolumn{1}{c|}{-} &
  \multicolumn{1}{c|}{daily} &
  \multicolumn{1}{c}{mean} \\ \cline{2-7} 
\multicolumn{1}{c|}{} &
  exercise &
  \multicolumn{1}{c|}{arbitrary} &
  \multicolumn{1}{c|}{count} &
  \multicolumn{1}{c|}{-} &
  \multicolumn{1}{c|}{daily} & 
  \multicolumn{1}{c}{zero} \\ \cline{2-7} 
\multicolumn{1}{c|}{} &
  exercise duration &
  \multicolumn{1}{c|}{arbitrary} &
  \multicolumn{1}{c|}{sum} &
  \multicolumn{1}{c|}{-} &
  \multicolumn{1}{c|}{daily} &
  \multicolumn{1}{c}{zero} \\ \cline{2-7} 
\multicolumn{1}{c|}{} &
  vo2max &
  \multicolumn{1}{c|}{2-day} &
  \multicolumn{1}{c|}{mean} &
  \multicolumn{1}{c|}{-} &
  \multicolumn{1}{c|}{*} &
  \multicolumn{1}{c}{*} \\ \cline{2-7} 
\multicolumn{1}{c|}{} &
  step goal &
  \multicolumn{1}{c|}{daily} &
  \multicolumn{1}{c|}{last} &
  \multicolumn{1}{c|}{-} &
  \multicolumn{1}{c|}{daily} &
  \multicolumn{1}{c}{zero} \\ \hline \hline
\multicolumn{1}{c|}{\multirow{4}{*}{Sleep}} &
  sleep duration &
  \multicolumn{1}{c|}{daily} &
  \multicolumn{1}{c|}{mean} &
  \multicolumn{1}{c|}{-} &
  \multicolumn{1}{c|}{forward} &
  \multicolumn{1}{c}{zero} \\ \cline{2-7} 
\multicolumn{1}{c|}{} &
  sleep points &
  \multicolumn{1}{c|}{daily} &
  \multicolumn{1}{c|}{mean} &
  \multicolumn{1}{c|}{-} &
  \multicolumn{1}{c|}{daily} &
  \multicolumn{1}{c}{mean} \\ \cline{2-7} 
\multicolumn{1}{c|}{} &
\begin{tabular}[c]{@{}l@{}} full/deep/light/rem \\ sleep breathing rate\end{tabular} &
  \multicolumn{1}{c|}{daily} &
  \multicolumn{1}{c|}{mean} &
  \multicolumn{1}{c|}{-} &
  \multicolumn{1}{c|}{*} &
  \multicolumn{1}{c}{*} \\ \cline{2-7} 
\multicolumn{1}{c|}{} &
  nightly temperature  &
  \multicolumn{1}{c|}{daily} &
  \multicolumn{1}{c|}{mean} &
  \multicolumn{1}{c|}{-} &
  \multicolumn{1}{c|}{*} &
  \multicolumn{1}{c}{*} \\ \hline \hline
\multicolumn{1}{c|}{\multirow{11}{*}{Health}} &
  water amount &
  \multicolumn{1}{c|}{arbitrary} &
  \multicolumn{1}{c|}{sum} &
  \multicolumn{1}{c|}{-} &
  \multicolumn{1}{c|}{daily} &
  \multicolumn{1}{c}{drop} \\ \cline{2-7} 
\multicolumn{1}{c|}{} &
   oxygen variation &
  \multicolumn{1}{c|}{2-minutes} &
  \multicolumn{1}{c|}{mean} &
  \multicolumn{1}{c|}{-} &
  \multicolumn{1}{c|}{*} &
  \multicolumn{1}{c}{*} \\ \cline{2-7} 
\multicolumn{1}{c|}{} &
  spo2 &
  \multicolumn{1}{c|}{daily} &
  \multicolumn{1}{c|}{mean} &
  \multicolumn{1}{c|}{-} &
  \multicolumn{1}{c|}{*} &
  \multicolumn{1}{c}{*} \\ \cline{2-7} 
\multicolumn{1}{c|}{} &
  ecg &
  \multicolumn{1}{c|}{arbitrary} &
  \multicolumn{1}{c|}{last} &
  \multicolumn{1}{c|}{*} &
  \multicolumn{1}{c|}{*} &
  \multicolumn{1}{c}{*} \\ \cline{2-7} 
\multicolumn{1}{c|}{} &
  nremhr &
  \multicolumn{1}{c|}{5-minutes} &
  \multicolumn{1}{c|}{-} &
  \multicolumn{1}{c|}{-} &
  \multicolumn{1}{c|}{*} &
  \multicolumn{1}{c}{*} \\ \cline{2-7} 
\multicolumn{1}{c|}{} &
  rmssd &
  \multicolumn{1}{c|}{5-minutes} &
  \multicolumn{1}{c|}{-} &
  \multicolumn{1}{c|}{-} &
  \multicolumn{1}{c|}{*} &
  \multicolumn{1}{c}{*} \\ \cline{2-7} 
\multicolumn{1}{c|}{} &
  heart rate alert &
  \multicolumn{1}{c|}{arbitrary} &
  \multicolumn{1}{c|}{last} &
  \multicolumn{1}{c|}{*} &
  \multicolumn{1}{c|}{*} &
  \multicolumn{1}{c}{*} \\ \cline{2-7} 
\multicolumn{1}{c|}{} &
  resting heart rate &
  \multicolumn{1}{c|}{4-day} &
  \multicolumn{1}{c|}{mean} &
  \multicolumn{1}{c|}{-} &
  \multicolumn{1}{c|}{*} &
  \multicolumn{1}{c}{*} \\ \cline{2-7} 
\multicolumn{1}{c|}{} &
  wrist temperature &
  \multicolumn{1}{c|}{3-minutes} &
  \multicolumn{1}{c|}{mean} &
  \multicolumn{1}{c|}{-} &
  \multicolumn{1}{c|}{*} &
  \multicolumn{1}{c}{*} \\ \cline{2-7} 
\multicolumn{1}{c|}{} &
  bpm &
  \multicolumn{1}{c|}{<1-minute} &
  \multicolumn{1}{c|}{mean} &
  \multicolumn{1}{c|}{-} &
  \multicolumn{1}{c|}{*} &
  \multicolumn{1}{c}{*} \\ \cline{2-7} 
\multicolumn{1}{c|}{} &
  calories &
  \multicolumn{1}{c|}{3-minutes} &
  \multicolumn{1}{c|}{sum} &
  \multicolumn{1}{c|}{-} &
  \multicolumn{1}{c|}{-} &
   \multicolumn{1}{c}{mean} \\ \hline \hline
\multicolumn{1}{c|}{\multirow{9}{*}{\begin{tabular}[c]{@{}c@{}}Mental\\ Health\end{tabular}}} &
  mood value &
  \multicolumn{1}{c|}{arbitrary} &
  \multicolumn{1}{c|}{mean} &
  \multicolumn{1}{c|}{-} &
  \multicolumn{1}{c|}{*} &
  \multicolumn{1}{c}{*} \\ \cline{2-7} 
\multicolumn{1}{c|}{} &
   scl avg &
  \multicolumn{1}{c|}{<1-minute} &
  \multicolumn{1}{c|}{mean} &
  \multicolumn{1}{c|}{-} &
  \multicolumn{1}{c|}{*} &
  \multicolumn{1}{c}{*} \\ \cline{2-7} 
\multicolumn{1}{c|}{} &
  mindfulness start hr &
  \multicolumn{1}{c|}{arbitrary} &
  \multicolumn{1}{c|}{mean} &
  \multicolumn{1}{c|}{-} &
  \multicolumn{1}{c|}{} &
  \multicolumn{1}{c}{*} \\ \cline{2-7} 
\multicolumn{1}{c|}{} &
  mindfulness end hr &
  \multicolumn{1}{c|}{arbitrary} &
  \multicolumn{1}{c|}{mean} &
  \multicolumn{1}{c|}{-} &
  \multicolumn{1}{c|}{} &
  \multicolumn{1}{c}{*} \\ \cline{2-7} 
\multicolumn{1}{c|}{} &
  stress score &
  \multicolumn{1}{c|}{daily} &
  \multicolumn{1}{c|}{mean} &
  \multicolumn{1}{c|}{-} &
  \multicolumn{1}{c|}{*} &
  \multicolumn{1}{c}{*} \\ \cline{2-7} 
\multicolumn{1}{c|}{} &
  stai stress &
  \multicolumn{1}{c|}{weekly} &
  \multicolumn{1}{c|}{last} &
  \multicolumn{1}{c|}{*} &
  \multicolumn{1}{c|}{*} &
  \multicolumn{1}{c}{*} \\ \cline{2-7} 
\multicolumn{1}{c|}{} &
  mood &
  \multicolumn{1}{c|}{thrice daily} &
  \multicolumn{1}{c|}{last} &
  \multicolumn{1}{c|}{*} &
  \multicolumn{1}{c|}{*} &
  \multicolumn{1}{c}{*} \\ \cline{2-7} 
\multicolumn{1}{c|}{} &
  exertion points &
  \multicolumn{1}{c|}{daily} &
  \multicolumn{1}{c|}{mean} &
  \multicolumn{1}{c|}{-} &
  \multicolumn{1}{c|}{daily} &
  \multicolumn{1}{c}{mean} \\ \cline{2-7} 
\multicolumn{1}{c|}{} &
   responsiveness points &
  \multicolumn{1}{c|}{daily} &
  \multicolumn{1}{c|}{mean} &
  \multicolumn{1}{c|}{-} &
  \multicolumn{1}{c|}{*} &
  \multicolumn{1}{c}{*} \\ \hline \hline
\multicolumn{1}{c|}{\multirow{3}{*}{Other}} &
  badges &
  \multicolumn{1}{c|}{arbitrary} &
  \multicolumn{1}{c|}{count} &
  \multicolumn{1}{c|}{-} &
  \multicolumn{1}{c|}{forward} &
  \multicolumn{1}{c}{zero} \\ \cline{2-7} 
\multicolumn{1}{c|}{} &
  place &
  \multicolumn{1}{c|}{thrice daily} &
  \multicolumn{1}{c|}{last} &
  \multicolumn{1}{c|}{one-hot} &
  \multicolumn{1}{c|}{backward} &
  \multicolumn{1}{c}{zero} \\ \cline{2-7} 
\multicolumn{1}{c|}{} &
  mindfulness goal &
  \multicolumn{1}{c|}{arbitrary} &
  \multicolumn{1}{c|}{last} &
  \multicolumn{1}{c|}{-} &
  \multicolumn{1}{c|}{daily} &
   \multicolumn{1}{c}{drop} \\ \hline \hline
\multicolumn{1}{c|}{\multirow{3}{*}{Demographics}} &
  gender &
  \multicolumn{1}{c|}{entry} &
  \multicolumn{1}{c|}{-} &
  \multicolumn{1}{c|}{*} &
  \multicolumn{1}{c|}{*} &
   \multicolumn{1}{c}{*} \\ \cline{2-7} 
\multicolumn{1}{l|}{} &
  age &
  \multicolumn{1}{c|}{entry} &
  \multicolumn{1}{c|}{-} &
  \multicolumn{1}{c|}{*} &
  \multicolumn{1}{c|}{*} &
   \multicolumn{1}{c}{*} \\ \cline{2-7} 
\multicolumn{1}{l|}{} &
  bmi &
  \multicolumn{1}{c|}{entry} &
  \multicolumn{1}{c|}{-} &
  \multicolumn{1}{c|}{*} &
  \multicolumn{1}{c|}{*} &
  \multicolumn{1}{c}{*} \\ \hline \hline
\multicolumn{1}{c|}{\multirow{5}{*}{Personality}} &
  extraversion  &
  \multicolumn{1}{c|}{entry} &
  \multicolumn{1}{c|}{-} &
  \multicolumn{1}{c|}{*} &
  \multicolumn{1}{c|}{*} &
   \multicolumn{1}{c}{*} \\ \cline{2-7} 
\multicolumn{1}{l|}{} &
  agreeableness  &
  \multicolumn{1}{c|}{entry} &
  \multicolumn{1}{c|}{-} &
  \multicolumn{1}{c|}{*} &
  \multicolumn{1}{c|}{*} &
   \multicolumn{1}{c}{*} \\ \cline{2-7} 
\multicolumn{1}{l|}{} &
   conscientiousness &
  \multicolumn{1}{c|}{entry} &
  \multicolumn{1}{c|}{-} &
  \multicolumn{1}{c|}{*} &
  \multicolumn{1}{c|}{*} &
   \multicolumn{1}{c}{*} \\ \cline{2-7} 
\multicolumn{1}{l|}{} &
  stability  &
  \multicolumn{1}{c|}{entry} &
  \multicolumn{1}{c|}{-} &
  \multicolumn{1}{c|}{*} &
  \multicolumn{1}{c|}{*} &
   \multicolumn{1}{c}{*} \\ \cline{2-7} 
\multicolumn{1}{l|}{} &
  intellect  &
  \multicolumn{1}{c|}{entry} &
  \multicolumn{1}{c|}{-} &
  \multicolumn{1}{c|}{*} &
  \multicolumn{1}{c|}{*} &
   \multicolumn{1}{c}{*} \\ \hline \hline
\multicolumn{1}{c|}{\multirow{14}{*}{Behavior}} &
  self-determination &
  \multicolumn{1}{c|}{entry/exit} &
  \multicolumn{1}{c|}{last} &
  \multicolumn{1}{c|}{*} &
  \multicolumn{1}{c|}{*} &
   \multicolumn{1}{c}{*} \\ \cline{2-7}
\multicolumn{1}{l|}{} &
  ttm stage &
  \multicolumn{1}{c|}{entry/exit} &
  \multicolumn{1}{c|}{last} &
  \multicolumn{1}{c|}{*} &
  \multicolumn{1}{c|}{*} &
   \multicolumn{1}{c}{*} \\ \cline{2-7}  
\multicolumn{1}{l|}{} &
  consciousness raising &
  \multicolumn{1}{c|}{entry/exit} &
  \multicolumn{1}{c|}{last} &
  \multicolumn{1}{c|}{ordinal} &
  \multicolumn{1}{c|}{periodic} &
   \multicolumn{1}{c}{mode} \\ \cline{2-7} 
\multicolumn{1}{l|}{} &
  dramatic relief &
  \multicolumn{1}{c|}{entry/exit} &
  \multicolumn{1}{c|}{last} &
  \multicolumn{1}{c|}{*} &
  \multicolumn{1}{c|}{*} &
   \multicolumn{1}{c}{*} \\ \cline{2-7}  
\multicolumn{1}{l|}{} &
  environmental reevaluation  &
  \multicolumn{1}{c|}{entry/exit} &
  \multicolumn{1}{c|}{last} &
  \multicolumn{1}{c|}{*} &
  \multicolumn{1}{c|}{*} &
   \multicolumn{1}{c}{*} \\ \cline{2-7}  
\multicolumn{1}{l|}{} &
  self-reevaluation &
  \multicolumn{1}{c|}{entry/exit} &
  \multicolumn{1}{c|}{last} &
  \multicolumn{1}{c|}{*} &
  \multicolumn{1}{c|}{*} &
   \multicolumn{1}{c}{*} \\ \cline{2-7}  
\multicolumn{1}{l|}{} &
  stimulus control &
  \multicolumn{1}{c|}{entry/exit} &
  \multicolumn{1}{c|}{last} &
  \multicolumn{1}{c|}{ordinal} &
  \multicolumn{1}{c|}{periodic} &
   \multicolumn{1}{c}{mode} \\ \cline{2-7} 
\multicolumn{1}{l|}{} &
  social liberation  &
  \multicolumn{1}{c|}{entry/exit} &
  \multicolumn{1}{c|}{last} &
  \multicolumn{1}{c|}{*} &
  \multicolumn{1}{c|}{*} &
   \multicolumn{1}{c}{*} \\ \cline{2-7}  
\multicolumn{1}{l|}{} &
  counter-conditioning &
  \multicolumn{1}{c|}{entry/exit} &
  \multicolumn{1}{c|}{last} &
  \multicolumn{1}{c|}{ordinal} &
  \multicolumn{1}{c|}{periodic} &
   \multicolumn{1}{c}{mode} \\ \cline{2-7} 
\multicolumn{1}{l|}{} &
  reinforcement management &
  \multicolumn{1}{c|}{entry/exit} &
  \multicolumn{1}{c|}{last} &
  \multicolumn{1}{c|}{*} &
  \multicolumn{1}{c|}{*} &
   \multicolumn{1}{c}{*} \\ \cline{2-7}  
\multicolumn{1}{l|}{} &
  self-liberation &
  \multicolumn{1}{c|}{entry/exit} &
  \multicolumn{1}{c|}{last} &
  \multicolumn{1}{c|}{*} &
  \multicolumn{1}{c|}{*} &
   \multicolumn{1}{c}{*} \\ \cline{2-7}  
\multicolumn{1}{l|}{} &
  helping relationships &
  \multicolumn{1}{c|}{entry/exit} &
  \multicolumn{1}{c|}{last} &
  \multicolumn{1}{c|}{ordinal} &
  \multicolumn{1}{c|}{periodic} &
   \multicolumn{1}{c}{mode} \\ \cline{2-7} 
\multicolumn{1}{l|}{} &
  negative affect score &
  \multicolumn{1}{c|}{weekly} &
  \multicolumn{1}{c|}{last} &
  \multicolumn{1}{c|}{*} &
  \multicolumn{1}{c|}{*} &
  \multicolumn{1}{c}{*} \\ \cline{2-7}
\multicolumn{1}{l|}{} &
  positive affect score &
  \multicolumn{1}{c|}{weekly} &
  \multicolumn{1}{c|}{last} &
  \multicolumn{1}{c|}{*} &
  \multicolumn{1}{c|}{*} &
  \multicolumn{1}{c}{*} \\ \hline
\end{tabular}
\end{table}

\subsection{Dataset Variants}\label{datasetvariants}

In section \ref{applicationdomain}, we mentioned that preprocessing and feature engineering produced six dataset variants for LifeSnaps, comprising three feature sets at both hourly and daily granularities. These variants will be used to explore how different feature categories affect the performance of foundational clustering algorithms. Details, including size and specific features, are provided in Tab. \ref{tab:featurevariants}. In summary, ``full'' and ``categories'' variants contain features from their respective categories, as shown in Tab. \ref{tab:preprocessingsteps}. The ``clean'' variant includes features with less than 60\% missing values before preprocessing, ensuring fewer imputed entries and potentially richer information; these features are: id, date, altitude, lightly active minutes, moderately active minutes, sedentary minutes, very active minutes, steps, minutes below zone 1, minutes in zones 1, 2, and 3, calories, consciousness-raising, counterconditioning, helping relationships, and stimulus control (grouped under ``set\_3'' in Tab. \ref{tab:featurevariants}).

\begin{table}[htp]
\scriptsize
\centering
\caption{The different feature variants of the LifeSnaps dataset.}
\label{tab:featurevariants}
\begin{tabular}{l|c|c|l}
\hline 
\textbf{Data variant} & \textbf{Rows} & \textbf{Columns} & \multicolumn{1}{c}{\textbf{Features}} \\ \hline \hline
hourly full & 159783 & 38 & physical activity, sleep, health, behavior, date, other \\ \hline
hourly categories & 159783 & 19 & physical activity, sleep, health \\ \hline
hourly clean & 159783 & 17 & set\_3 \\ \hline
daily full & 8292 & 36 & physical activity, sleep, health, behavior, date, other \\ \hline
daily categories  & 8292 & 19 & physical activity, sleep, health  \\ \hline
daily clean & 8292 & 17 & set\_3 \\ \hline
\end{tabular}
\end{table}

\section{The clustering analysis}\label{appandixB}
In section \ref{thesaurus}, we highlighted the k-means as the best-performing clustering algorithm on the hourly granularity on the ``categories'' feature subset. In Tab. \ref{tab:clusteringperformance} we present the performance of each clustering algorithm (under its optimized hyperparameter setup) for all six different dataset variants, using multiple evaluation metrics. In terms of metrics, the Silhouette Score ranges from -1 to 1, indicating how well-defined clusters are. The Davies-Bouldin Index (DBI) measures cluster separation, with values close to 0 indicating dense clustering. The Calinski-Harabasz Index (CHI) focuses on variance, where higher values signify better clustering. The Dunn Index assesses the smallest distance between clusters relative to the largest within-cluster distance, with higher values indicating better clustering. The PBM Index evaluates cluster compactness and separation, with higher values indicating high-dense clustering. The Xie-Beni Index, used in fuzzy clustering, measures membership degrees, with lower values indicating optimal clustering. The symbol '-' indicates where an algorithm failed to run due to computational complexity.

\begin{table*}[htp]
\centering
\scriptsize
\caption{The performance of clustering algorithms across the different dataset variants measured with various metrics.}
\label{tab:clusteringperformance}
\begin{tabular}{c|cl|c|c|c|c|c|c|c}
\hline 
\multicolumn{1}{l}{} &
  \multicolumn{1}{l}{} &
   & \multicolumn{7}{c}{\textbf{Evaluation metrics}} \\ \hline
\textbf{\begin{tabular}[c]{@{}c@{}}Clustering\\ algorithm\end{tabular}} &
  \multicolumn{2}{c|}{\textbf{\begin{tabular}[c]{@{}c@{}}Dataset\\ variant\end{tabular}}} &
  \textbf{Silhouette} &
  \textbf{DBI} &
  \textbf{CHI} &
  \textbf{Dunn} &
  \textbf{PBM} &
  \textbf{Xie-Beni} &
  \textbf{k} \\ \hline \hline
\multirow{6}{*}{\textcolor{teal}{\textbf{k-means}}}  & \multicolumn{1}{c|}{\multirow{3}{*}{\textcolor{teal}{\textbf{hourly}}}} & full       & 0.15  & 1.96 & 21075.94 & 0.45 & 0.20 & 1.03  & 4   \\ \cline{3-10} 
                                   & \multicolumn{1}{c|}{}                        & \textcolor{teal}{\textbf{categories}} & \textcolor{teal}{\textbf{0.35}}  & 1.18 & 99605.72 & 0.37 & 0.02 & 0.59  & 2   \\ \cline{3-10} 
                                   & \multicolumn{1}{c|}{}                        & clean      & 0.32  & 1.33 & 82890.54 & 0.58 & 0.12 & 0.66  & 2   \\ \cline{2-10} 
                                   & \multicolumn{1}{c|}{\multirow{3}{*}{daily}}  & full       & 0.15  & 1.95 & 1127.23  & 0.45 & 0.20 & 1.03  & 4   \\ \cline{3-10} 
                                   & \multicolumn{1}{c|}{}                        & categories & 0.32  & 1.25 & 4537.61  & 0.41 & 0.02 & 0.62  & 2   \\ \cline{3-10} 
                                   & \multicolumn{1}{c|}{}                        & clean      & 0.29  & 1.40 & 3864.86  & 0.54 & 0.09 & 0.70  & 2   \\ \hline
\multirow{6}{*}{\textbf{spectral}} & \multicolumn{1}{c|}{\multirow{3}{*}{hourly}} & full       & -     & -    & -        & -    & -    & -     & -   \\ \cline{3-10}
                                   & \multicolumn{1}{c|}{}                        & categories & -     & -    & -        & -    & -    & -     & -   \\ \cline{3-10}
                                   & \multicolumn{1}{c|}{}                        & clean      & -     & -    & -        & -    & -    & -     & -   \\ \cline{2-10}
                                   & \multicolumn{1}{c|}{\multirow{3}{*}{daily}}  & full       & 0.14  & 2.43 & 1330.65  & 0.40 & 0.19 & 1.21  & 2   \\ \cline{3-10}
                                   & \multicolumn{1}{c|}{}                        & categories & 0.32  & 1.24 & 4532.05  & 0.41 & 0.02 & 0.62  & 2   \\ \cline{3-10}
                                   & \multicolumn{1}{c|}{}                        & clean      & 0.29  & 1.41 & 3825.80  & 0.54 & 0.10 & 0.70  & 2   \\ \hline
\multirow{6}{*}{\textbf{fuzzy}}    & \multicolumn{1}{c|}{\multirow{3}{*}{hourly}} & full       & 0.14  & 2.43 & 26084.93 & 0.43 & 0.19 & 1.21  & 2   \\ \cline{3-10} 
                                   & \multicolumn{1}{c|}{}                        & categories & 0.34  & 1.18 & 99562.77 & 0.37 & 0.02 & 0.59  & 2   \\ \cline{3-10}
                                   & \multicolumn{1}{c|}{}                        & clean      & 0.32  & 1.33 & 82779.53 & 0.58 & 0.12 & 0.66  & 2   \\ \cline{2-10}
                                   & \multicolumn{1}{c|}{\multirow{3}{*}{daily}}  & full       & 0.14  & 2.41 & 1352.69  & 0.41 & 0.16 & 1.20  & 2   \\ \cline{3-10}
                                   & \multicolumn{1}{c|}{}                        & categories & 0.32  & 1.25 & 4534.93  & 0.41 & 0.02 & 0.62  & 2   \\ \cline{3-10}
                                   & \multicolumn{1}{c|}{}                        & clean      & 0.29  & 1.41 & 3854.72  & 0.54 & 0.10 & 0.70  & 2   \\ \cline{1-10}
\multirow{6}{*}{\textbf{DBSCAN}}   & \multicolumn{1}{c|}{\multirow{3}{*}{hourly}} & full       & -0.23 & 1.93 & 502.02   & 0.21 & 0.04 & 2.57  & 11  \\ \cline{3-10}
                                   & \multicolumn{1}{c|}{}                        & categories & -0.09 & 1.19 & 1768.89  & 0.05 & 0.00 & 3.28  & 78  \\ \cline{3-10}
                                   & \multicolumn{1}{c|}{}                        & clean      & 0.10  & 1.22 & 2780.53  & 0.10 & 0.00 & 1.94  & 147 \\ \cline{2-10}
                                   & \multicolumn{1}{c|}{\multirow{3}{*}{daily}}  & full       & 0.03  & 2.50 & 408.95   & 0.32 & 0.11 & 1.52  & 6   \\ \cline{3-10}
                                   & \multicolumn{1}{c|}{}                        & categories & 0.12  & 1.79 & 280.80   & 0.31 & 0.02 & 1.11  & 4   \\ \cline{3-10} 
                                   & \multicolumn{1}{c|}{}                        & clean      & 0.22  & 1.40 & 370.93   & 0.17 & 0.00 & 1.26  & 46  \\ \hline
\multirow{6}{*}{\textbf{HDBSCAN}}  & \multicolumn{1}{c|}{\multirow{3}{*}{hourly}} & full       & 0.09  & 3.34 & 10490.41 & 0.29 & 0.12 & 1.73  & 5   \\ \cline{3-10} 
                                   & \multicolumn{1}{c|}{}                        & categories & -0.08 & 1.48 & 7932.35  & 0.15 & 0.00 & 1.55  & 6   \\ \cline{3-10}
                                   & \multicolumn{1}{c|}{}                        & clean      & 0.29  & 1.23 & 8360.53  & 0.01 & 0.00 & 10.04 & 66  \\ \cline{2-10}
                                   & \multicolumn{1}{c|}{\multirow{3}{*}{daily}}  & full       & 0.10  & 2.72 & 540.78   & 0.37 & 0.08 & 1.30  & 6   \\ \cline{3-10}
                                   & \multicolumn{1}{c|}{}                        & categories & 0.06  & 2.39 & 1313.29  & 0.17 & 0.01 & 1.55  & 3   \\ \cline{3-10}
                                   & \multicolumn{1}{c|}{}                        & clean      & 0.25  & 1.40 & 513.03   & 0.08 & 0.00 & 2.22  & 45  \\ \hline
\multirow{6}{*}{\textbf{ROBP}}     & \multicolumn{1}{c|}{\multirow{3}{*}{hourly}} & full       & -     & -    & -        & -    & -    & -     & -   \\ \cline{3-10} 
                                   & \multicolumn{1}{c|}{}                        & categories & -     & -    & -        & -    & -    & -     & -   \\ \cline{3-10}
                                   & \multicolumn{1}{c|}{}                        & clean      & -     & -    & -        & -    & -    & -     & -   \\ \cline{2-10}
                                   & \multicolumn{1}{c|}{\multirow{3}{*}{daily}}  & full       & -0.02 & 2.13 & 195.55   & 0.23 & 0.00 & 1.76  & 32  \\ \cline{3-10}
                                   & \multicolumn{1}{c|}{}                        & categories & -0.16 & 1.97 & 252.63   & 0.08 & 0.00 & 3.36  & 11  \\ \cline{3-10}
                                   & \multicolumn{1}{c|}{}                        & clean      & 0.21  & 1.54 & 491.03   & 0.05 & 0.00 & 3.58  & 50  \\ \cline{1-10}
\end{tabular}
\end{table*}

\begin{credits}
\subsubsection{\ackname} Funded by the European Union (GILL, 101094812). Views and opinions expressed are, however, those of the author(s) only and do not necessarily reflect those of the European Union or the European Research Executive Agency (REA). Neither the European Union nor the European Research Executive Agency can be held responsible for them. This project is co-funded by UK Research and Innovation (UKRI) under the UK government’s Horizon Europe funding guarantee [grant number 10049511]. Partially funded by the European Union grants under the Marie Skłodowska-Curie Grant Agreements No: 101169474 (AlignAI).

\end{credits}
%
%
%
\bibliographystyle{splncs04}
\bibliography{bibliography}

\end{document}